%% file: main.tex
\documentclass{ecai} 

\usepackage{latexsym}
\usepackage{amssymb}
\usepackage{amsmath}
\usepackage{amsthm}
\usepackage{booktabs}
\usepackage{enumitem}
\usepackage{graphicx}
\usepackage{color}
\usepackage{minted}

\usepackage{algpseudocode}

\usepackage{url}

\newtheorem{theorem}{Theorem}
\newtheorem{lemma}[theorem]{Lemma}

\newcommand{\BibTeX}{B\kern-.05em{\sc i\kern-.025em b}\kern-.08em\TeX}

\usepackage{framed}
\usepackage{amssymb,amsfonts,amsmath,amsthm} 
\usepackage{hyperref}
\usepackage{url}
\usepackage{float}
\usepackage{subcaption} 
\usepackage{enumitem,kantlipsum}
\usepackage{color}
\usepackage{xcolor} 

\usepackage{graphicx}
\usepackage{booktabs} 

\input{math_commands.tex}

\usepackage[capitalize,noabbrev]{cleveref}

\usepackage[ruled]{algorithm2e}
\SetKwInput{KwInput}{Input}                
\SetKwInput{KwOutput}{Output}              

\usepackage[textsize=tiny]{todonotes}

\begin{document}


\begin{frontmatter}


\paperid{1171} 

\title{A Non-Adversarial Approach \\ to Idempotent Generative Modelling}


\author[1]{\fnms{Mohammed}~\snm{Al-Jaff}\thanks{Corresponding Author. Email: mohamaj@kth.se.}}
\author[1]{\fnms{Giovanni Luca}~\snm{Marchetti}}
\author[1]{\fnms{Michael C}~\snm{Welle}}
\author[1]{\fnms{Jens}~\snm{Lundell}}
\author[2]{\fnms{Mats G.}~\snm{Gustafsson}}
\author[1]{\fnms{Gustav Eje}~\snm{Henter}}
\author[1]{\fnms{Hossein}~\snm{Azizpour}}
\author[1]{\fnms{Danica}~\snm{Kragic}}

\address[1]{KTH Royal Institute of Technology}
\address[2]{Uppsala University}

\begin{abstract}
    Idempotent Generative Networks (IGNs) are deep generative models that also function as local data manifold projectors, mapping arbitrary inputs back onto the manifold. 
    They are trained to act as identity operators on the data and as idempotent operators off the data manifold. 
    However, IGNs suffer from mode collapse, mode dropping, and training instability due to their objectives, which contain adversarial components and can cause the model to cover the data manifold only partially --  an issue shared with generative adversarial networks. 
    We introduce Non-Adversarial Idempotent Generative Networks (NAIGNs) to address these issues. Our loss function combines reconstruction with the non-adversarial generative objective of Implicit Maximum Likelihood Estimation (IMLE). This improves on IGN's ability to restore corrupted data and generate new samples that closely match the data distribution. 
    We moreover demonstrate that NAIGNs implicitly learn the distance field to the data manifold, as well as an energy-based model. 
\end{abstract}

\end{frontmatter}

\section{Introduction and Related Work}
 Generative modeling is one of the fundamental tasks in modern machine learning \citep{dgm}. Contemporary deep generative models are characterized by their ability to transform random noise into data-like outputs, enabling the synthesis of new, realistic samples across various domains, e.g., image generation \citep{zhang2023text}, text synthesis \citep{li2024pre}, and 3D object modeling \citep{li2024advances}. Models such as Generative Adversarial Networks (GANs; \citep{goodfellow2020generative}), Variational Autoencoders (VAEs; \citep{kingma2013auto, rezende2014stochastic}), and diffusion models \citep{ho2020denoising} have demonstrated remarkable success in capturing complex data distributions and producing high-fidelity samples.

\begin{figure}[t!]
    \centering
    \includegraphics[width=1.\linewidth]{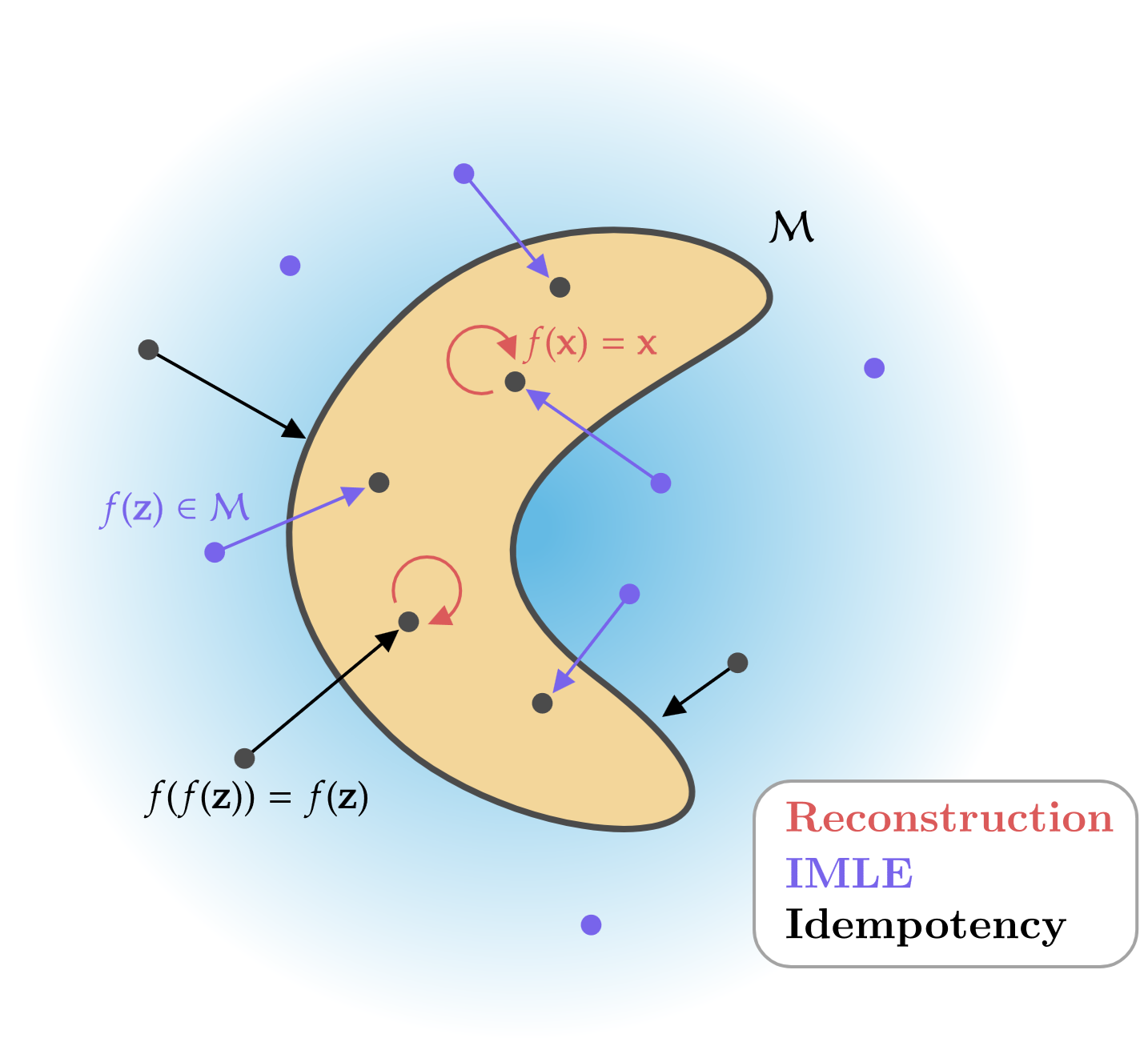}
    \caption{ NAIGNs $f$ are trained to fix points ($f(\mathbf{x})~=~\mathbf{x}$) on the data manifold $\mathcal{M}$ and to  map arbitrary points to $\mathcal{M}$ ($f(\mathbf{x}) \in \mathcal{M}$) via IMLE, which in turn imply idempotency ($f(f(\mathbf{x}))=f(\mathbf{x})$).} 
    \label{fig:firstpage}
    \vspace{2em}
\end{figure}

Another fundamental task in machine learning is manifold learning and projection \citep{izenman2012introduction}, consisting of mapping corrupted or out-of-distribution data points back onto the data manifold. The latter includes tasks related to image-to-image translation and domain adaptation ~\citep{cyclegan, da_cyclegan}. 
{The former represents the support of the underlying data distribution.} 
This is essential in applications requiring reconstruction, restoration, and robustness, including inverse problems such as denoising, deblurring, data imputation, and adversarial purification~\citep{nie2022diffusion, peng, zhu2023denoising}. 

        
\begin{figure*}[t!]
    \centering
    \includegraphics[width=1.\linewidth, trim={0pt 0pt 3pt 70pt}, clip]{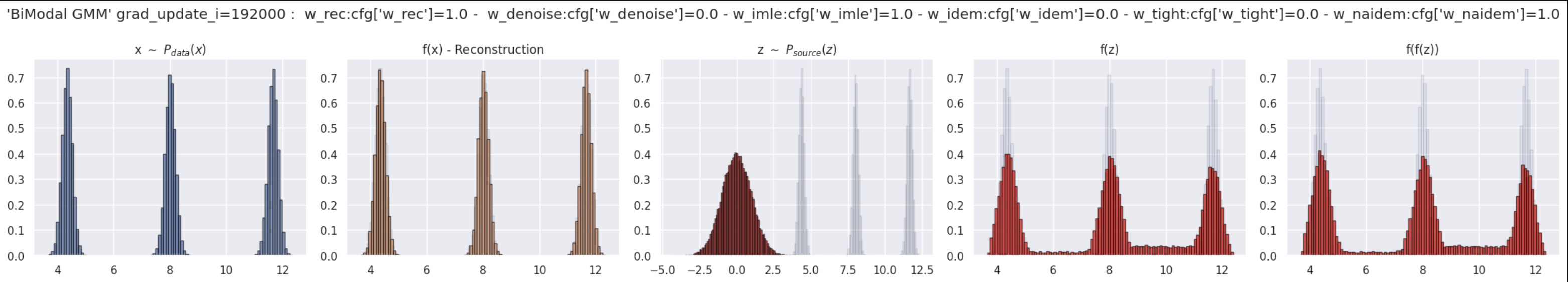}
    \includegraphics[width=1.\linewidth, trim={0pt 0pt 0pt 70pt}, clip]{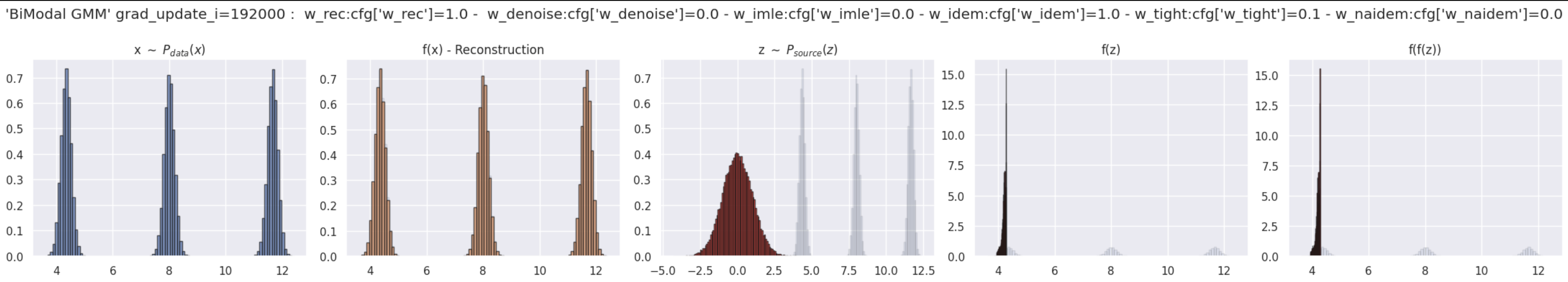}
    \begin{picture}(0,0)
        \put(-205,157){$p_{\text{data}}(\mathbf{x})$}
        \put(-105,157){$f_\theta(\mathbf{\mathbf{x}})$}
        \put(-5,157){$p_{Z}(\mathbf{z})$}
        \put(90,157){$f_\theta(\mathbf{z})$}
        \put(180,157){$f_\theta(f_\theta(\mathbf{z}))$}
    \end{picture}
    \caption{Comparison between NAIGN (top row) and IGN (bottom row) trained on a simple tri-modal one-dimensional distribution. Our proposed method, NAIGN, is better at mitigating mode collapse and mode dropping issues that IGN is susceptible to. For reference, the light gray histograms in the three rightmost columns are the target distribution from the first column.}
    \label{fig:1dexp}
    \vspace{1em}
\end{figure*}
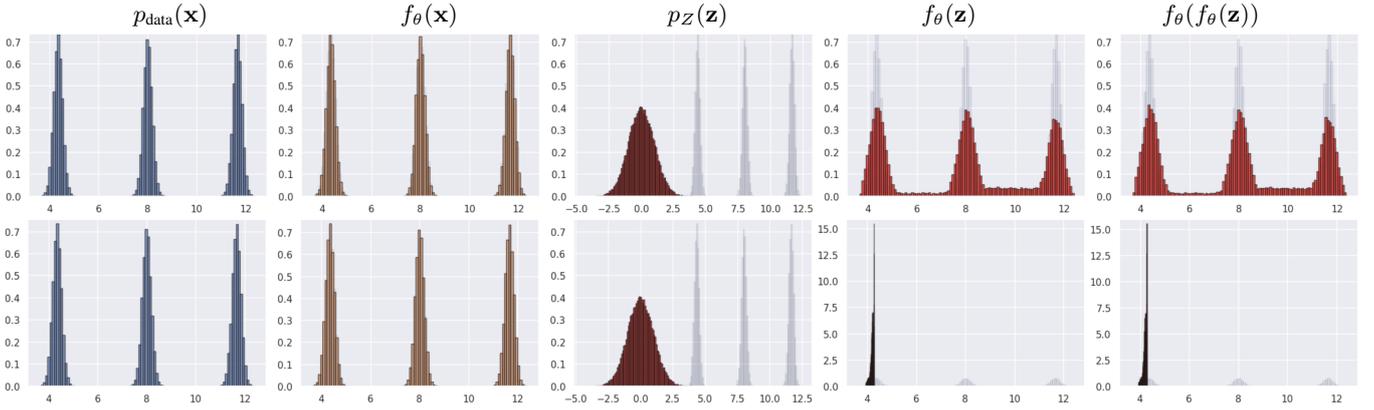

While generative modeling and manifold projection have traditionally been treated as separate tasks, recent advancements have sought to unify these capabilities within a single framework {\citep{gaga}}. Models that can both generate realistic samples and project onto the data manifold hold significant promise for tasks that simultaneously require synthesis and manifold distance estimation. 
Idempotent Generative Networks (IGNs; \citep{shocher2023idempotent}) represent a step in this direction, aiming to integrate these functionalities.

{
Formally, let \( f: \mathbb{R}^D \to \mathbb{R}^D \) denote a neural network, let \( \mathbf{z} \in \mathbb{R}^D \) be an arbitrary point in ambient space, and let \( \mathbf{x} \in \mathcal{M} \subset \mathbb{R}^D \) denote a point on the data manifold. IGNs are defined by their ability to act as identity operators on manifold points, i.e., \( f(\mathbf{x}) = \mathbf{x} \) when \( \mathbf{x} \in \mathcal{M} \), and as idempotent operators on the ambient space, satisfying \( f(f(\mathbf{z})) = f(\mathbf{z}) \). This dual property enables generation via projection of noise onto the data manifold.
}
        

In the case of IGNs, the generation process happens within a single (-or few) model forward passes. This is in contrast to diffusion-based models, where multiple passes are performed to incrementally map noise samples to the data manifold. However, {as we will show,} IGNs face challenges in mode collapse, mode dropping, and training instability due to adversarial components in their objective.
        
In this work, we aim to improve and enhance IGNs by addressing their issues. We propose Non-Adversarial Idempotent Generative Networks (NAIGN), introducing an alternative loss formulation that eliminates adversarial components. Our objective consists of a reconstruction loss term and an Implicit Maximum Likelihood Estimation (IMLE) term \citep{imle}, which together act as a generative and fixed-point loss -- see Figure \ref{fig:firstpage}. Our approach not only stabilizes training dynamics, but also enhances the model's ability to approximate the local distance field of the data manifold. 
        
 We empirically validate NAIGN and compare it with IGN on low-dimensional synthetic datasets, as well as on the MNIST image dataset. We investigate idempotency, stability, as well as generation and restoration capabilities. Furthermore, we showcase the performance of NAIGN as a manifold projector with applications to density estimation. The Python code for NAIGN and all supplementary material available at \url{https://github.com/MohammedAlJaff/naign} \citep{naign2025supp}

\section{Background}\label{sec:backgrnd}
        \subsection{Notation}
         Let \(p_{\text{data}}\) be an unknown distribution over an ambient space $\mathbb{R}^D$. Data is sampled i.i.d. from this distribution. In order to formalize the \emph{data manifold}, we assume that \(p_{\text{data}}\) is concentrated around a manifold \(\mathcal{M} \subset \mathbb{R}^D\). Let \(p_Z\) be a source distribution, e.g., standard normal, from which we sample \(\mathbf{z} \in \mathbb{R}^D\).
        Consider a neural network \(f_{\theta}: \mathbb{R}^D \rightarrow \mathbb{R}^D\) with parameters $\theta$.
        We define the \emph{drift} of the network as the distance between the input and the output:  $\delta_{\theta}(\mathbf{z}):= d(\mathbf{z}, f_{\theta}(\mathbf{z}))$, $\mathbf{z} \in \mathbb{R}^D$. Here, $d$ is some distance function, such as the Euclidean metric. To quantify how far any point \(\mathbf{z} \in \mathbb{R}^D\) is from the data manifold, we define the manifold \emph{distance field} as
        \begin{equation}\label{eq:distfield}
            d_{\mathcal{M}}(\mathbf{z}) = \min_{\mathbf{m} \in \mathcal{M}} d\left(\mathbf{z}, \mathbf{m}\right).
        \end{equation}
        This function measures the shortest distance from \(\mathbf{z}\) to any point on \(\mathcal{M}\), capturing the notion of proximity to the data manifold.

    \subsection{Idempotent Generative Networks}\label{sec:idemgen}
        This section reviews IGNs, which were introduced by \citet{shocher2023idempotent}. Two specific properties characterize these deep generative models. First, an ideal IGN fixes points on the data manifold. Formally, for \(\mathbf{x} \in \mathcal{M} \subset \mathbb{R}^D\) we have:
        \begin{equation}
            f_{\theta}(\mathbf{x}) = \mathbf{x}.
            \end{equation}
        Second, IGNs should be \emph{idempotent}, meaning that applying the model twice to any point in the ambient space is equivalent to applying it once. Formally, for \(\mathbf{z} \in \mathbb{R}^D\) we have:
        \begin{equation}
            f_\theta(f_{\theta}(\mathbf{z})) = f_\theta(\mathbf{z}).
       \end{equation}
    While idempotency and reconstruction are desirable properties, they do not ensure full coverage of $\mathcal{M}$, and even allow for trivial solutions, such as the identity function $f_\theta(\mathbf{z}) = \mathbf{z}$. This raises the need for careful loss design. \citet{shocher2023idempotent} propose  to circumvent these issues via a combination of three specific weighted loss functions:
        \begin{equation}
            \mathcal{L}_{\text{IGN}}(\theta) = w_{\text{rec}} \mathcal{L}_{\text{rec}}(\theta) + w_{\text{idem}} \mathcal{L}_{\text{idem}}(\theta) + w_{\text{tight}} \mathcal{L}_{\text{tight}}(\theta).
        \end{equation}
       The above loss terms are defined as follows.
        \begin{itemize}[wide, labelindent=0pt]
            \item \textbf{Reconstruction Loss}:
               \begin{equation}\label{eq:recloss}
               \mathcal{L}_{\text{rec}}(\theta) = \mathbb{E}_{\mathbf{x} \sim p_{\text{data}}} \left[ d\left( \mathbf{x}, f_{\theta}(\mathbf{x}) \right) \right],
                \end{equation}
               where \( d(\cdot, \cdot) \) is a distance function, e.g., Euclidean distance. This loss ensures that the IGN acts as the identity function on data manifold points.

            \item \textbf{Idempotency Loss}:
               \begin{equation}
               \mathcal{L}_{\text{idem}}(\theta) = \mathbb{E}_{\mathbf{z} \sim p_{Z}} \left[ d\left( f_{\theta}(\mathbf{z}), f_{\theta_{\bot}} \left( f_{\theta}(\mathbf{z}) \right) \right) \right],
                \end{equation}
                where \( \theta_{\bot} \) indicates that gradients are not propagated through \( f_{\theta_{\bot}} \) (i.e., stop-gradient operation is applied). This loss encourages the model to be idempotent. 

            \item \textbf{Tightness Loss}:
                \begin{equation}
               \mathcal{L}_{\text{tight}}(\theta) = -\mathbb{E}_{\mathbf{z} \sim p_{Z}} \left[ d\left( f_{\theta_{\bot}}(\mathbf{z}), f_{\theta}(f_{\theta_{\bot}}(\mathbf{z})) \right) \right].
               \end{equation}
               Here, the negative sign indicates that this loss term encourages the outputs of \( f_{\theta} \) to be as far as possible when applied iteratively.
        \end{itemize}

        A crucial aspect of the IGN's objective is the interplay between the idempotency loss \( \mathcal{L}_{\text{idem}} \) and the tightness loss \( \mathcal{L}_{\text{tight}} \). These two loss terms share a similar expression, but with opposite signs and with different gradient stopping protocols, i.e., they are \emph{adversarial}. 
        While $\mathcal{L}_{\text{idem}}$ forces idempotency, its adversarial counterpart $\mathcal{L}_{\text{tight}}$ prevents the model from collapsing to trivial idempotent solutions, such as the identity function. The stop-gradient operations result in a bootstrapped learning dynamics, where the two adversarial losses enhance each other iteratively throughout training,  acting as reciprocal fixed targets. \citet{shocher2023idempotent} demonstrated that under ideal conditions, minimizing \( \mathcal{L}_{\text{IGN}}(\theta) \) enables the model to learn the true data distribution, showing that the adversarial dynamics in IGN leads it to capture the data manifold accurately.  

        However, adversarial dynamics suffer from well-known issues \citep{gan}. First, they introduce instabilities, resulting in oscillations during training. Second, they typically lead to mode dropping and mode collapse. Mode dropping occurs when the model fails to cover certain regions (modes) of the data manifold. Mode collapse happens when the generator model focuses on certain regions of the data manifold, where most of its outputs get concentrated. We illustrate these phenomena in Figure \ref{fig:1dexp} (bottom row). Therefore, practical training of IGNs requires adjustments to be effective. These include carefully weighting each loss component -- sometimes even dynamically -- or altering the gradient flow. These practical considerations emphasize that, despite the sound theoretical foundation of IGNs, careful implementation and hyperparameter tuning are essential for achieving satisfactory results. In this work, we aim to design an idempotent generative model without any adversarial component, preventing these issues completely. 
        
        \subsection{Implicit Probabilistic Generative Models} \label{implicit}
        Implicit probabilistic generative models aim to capture an unknown target distribution \( p_{\text{data}} \) using a parameterized generator function $ f_{\theta}: \mathbb{R}^{d} \rightarrow \mathbb{R}^{D}$ in the general case where \( d \leq D \). These models employ a two-step generation process. First,  a latent variable \( \mathbf{z} \in \mathbb{R}^{d} \) is sampled from a simple source distribution \( p_{Z} \), such as \( \mathcal{N}(\mathbf{0}, I) \). Second, the generator transforms this latent variable into a data-like sample \( f_{\theta}(\mathbf{z}) \). 
        The resulting distribution of the outputs of the network is called the push-froward of $p_{Z}$ through $f_{\theta}$, and is denoted as
        \( p_{\theta} := [f_{\theta}]_{\sharp} p_{Z} \). The goal is for \( p_{\theta} \) to resemble \( p_{\text{data}} \). Such models are often referred to as \textit{push-forward generative models} \citep{salmona2022can}.
        
        Maximizing the likelihood of the training data is the ideal objective in probabilistic generative models. However, the likelihood function of the generated distribution $p_\theta$ is often intractable. While adversarial approaches like GANs \citep{gan} circumvent this issue, they come with major disadvantages, as discussed above. To overcome these limitations, \citet{imle} propose Implicit Maximum Likelihood Estimation (IMLE), a non-adversarial method that ensures better mode coverage. IMLE encourages the model to match the real data distribution by ensuring that each data point has at least one generated sample nearby. Let $\mathbf{x}_1, \ldots, \mathbf{x}_N$ be (a batch of) training data, and \( \mathbf{z}_1, \ldots, \mathbf{z}_M \) be independent samples from \( p_{Z} \), where \( M \) is typically significantly larger than $N$. The IMLE objective is:
        \begin{equation}\label{eq:imle}
        \mathcal{L}_{\text{IMLE}}(\theta) = \mathbb{E}_{ \mathbf{z}_1, \ldots, \mathbf{z}_M \sim p_{Z}} \left[ \sum_{i=1}^{N} \min_{j \in [M]} d\left(\mathbf{x}_i, f_{\theta}(\mathbf{z}_j)\right) \right],
        \end{equation}
        where \( d(\cdot, \cdot) \) is a metric, typically the Euclidean distance. \citet{imle} show that under reasonable conditions, IMLE approximately optimizes the maximum likelihood estimation objective, meaning that minimizing Equation \ref{eq:imle} is equivalent to maximizing \( \mathbb{E}_{\mathbf{x} \sim p_{\text{data}}} \left[ \log p_{\theta}(\mathbf{x}) \right] \). Moreover, the IMLE objective coincides with the one-sided Chamfer distance between the real dataset and the generated set. Given two finite sets \( A, B \subset \mathbb{R}^D \), the Chamfer distance is defined as:
        \begin{equation}
        d_{\text{Chmf}}(A, B) = \sum_{a \in A} \min_{b \in B} d(a, b).
        \end{equation}
        By minimizing this loss, IMLE ensures that the generated samples adequately cover the real data distribution, providing a remedy for both mode collapse and mode dropping.

\section{Method}
    \subsection{Non-Adversarial Idempotent Generative Networks}\label{sec:naignmeth}
        The main goals of idempotent generative models are twofold: 
        1) to be \emph{generative}, i.e., capable of mapping points from a source distribution, e.g., Gaussian noise, to the data manifold, and 2) to act as a \emph{restorer}, i.e., to take off-manifold data and project it back to the manifold in one or very few iterations. We propose an alternative approach for idempotent generative modeling that aims to both enhance generative capabilities and act as a manifold projector locally while mitigating issues -- such as mode collapse and the need for careful tuning of the weights -- that are caused by the adversarial loss in IGNs. 
        
        Our method is grounded in the observation that idempotency naturally arises in models satisfying two simple mapping conditions, which can be enforced through simple, straightforward loss functions. We aim to design a model $ f_\theta \colon \mathbb{R}^D  \rightarrow \mathbb{R}^D $ satisfying the following two conditions:
        \begin{itemize}[wide, labelindent=0pt]
             \item \textbf{Fixed point on data:} The model acts as the identity on data manifold points. Formally, $f_\theta(\mathbf{x}) = \mathbf{x}$ for all $\mathbf{x} \in \mathcal{M}$, where \( \mathcal{M} \subset \mathbb{R}^D \) denotes the data manifold.
             \item \textbf{Push-forward generative:} The model maps points in ambient space to the data manifold. Formally, $f_{\theta}(\mathbb{R}^D) = \mathcal{M}$. This is the geometric analogue of the probabilistic condition \( p_{\theta} = [f_{\theta}]_{\sharp} p_{Z} \) characterizing push-forward generative models. 
         \end{itemize}
         
         The above conditions imply that any input from the data manifold remains unchanged by $f_\theta$, while inputs from elsewhere are mapped onto the manifold. This naturally leads to idempotency, as shown below.  

        \begin{lemma}\label{lemm:idempt}
        If $f_\theta(\mathbf{x}) = \mathbf{x}$ for all $\mathbf{x} \in \mathcal{M}$ and $f_\theta(\mathbb{R}^D) = \mathcal{M}$, then $f_\theta$ is idempotent, i.e., $f_\theta(f_\theta(\mathbf{z})) = f_\theta(\mathbf{z})$ for all $\mathbf{z} \in \mathbb{R}^D$.  
         \end{lemma}
         \begin{proof}
            For $\mathbf{z} \in \mathbb{R}^D$, $f_\theta(\mathbf{z}) \in \mathcal{M}$ by the push-forward generative condition. But then the fixed point condition implies that $f_\theta(f_\theta(\mathbf{z})) = f_\theta(\mathbf{z})$, as desired. 
         \end{proof}
    Note that these conditions are stricter than those defined for IGNs, and, importantly, the trivial or collapsed idempotent solutions, which are issues for IGNs, will not satisfy the above conditions.

    Based on the above arguments, our proposed \textbf{N}on-\textbf{A}dverserial \textbf{I}dempotent \textbf{G}enerative \textbf{N}etworks (NAIGNs, pronounced `9') optimize the following loss:
    \begin{equation}\label{eq:ourloss}
       \Ls_{\text{rec}}(\theta) + \Ls_{\text{IMLE}}(\theta). 
    \end{equation}
   %
    The loss terms, which are defined in Section \ref{sec:idemgen}, are motivated as follows: 
    \begin{itemize}[wide, labelindent=0pt]
        \item $\Ls_{\text{rec}}$ ensures that the model fixes points on the data manifold.   
        \item  $\Ls_{\text{IMLE}}$ encourages the model to generate outputs that are on the data manifold or, equivalently, to produce a push-forward distribution \([f_{\theta}]_{\sharp} p_{Z}\) that matches \( p_{\text{data}} \). Via IMLE, we explicitly guide the model to learn the data distribution without adversarial training, enforcing mode coverage. 
        
        \end{itemize}

    While Lemma~\ref{lemm:idempt} holds for a perfect generator and fixed-point model, real-world models and data are far from ideal. Empirically, we find that explicitly including a modified idempotent term, $\Ls_{\text{idem}}$, provides better control over the trade-off between avoiding off-manifold generation and mode collapse (see Figure~\ref{fig:1dexp}). This term applies the idempotency loss a second time on the output; see the supplementary material \citep{naign2025supp} for details. With the additional idempotency term, our final loss becomes
    \begin{equation}
        \Ls_{\text{NAIGN}}(\theta) = \Ls_{\text{rec}}(\theta) + \Ls_{\text{IMLE}}(\theta) + \Ls_{\text{idem}}(\theta).
    \end{equation}
    
    Algorithm~\ref{alg:naign_training_concise} shows the pseudo-code for a single NAIGN training iteration. A PyTorch-based minimal version is provided in the supplementary material \citep{naign2025supp}.

\begin{algorithm}
\caption{NAIGN Training Step}
\label{alg:naign_training_concise}
\begin{algorithmic}[1]
\Require
    \Statex $f_{\theta}$: Model with parameters $\theta$
    \Statex $\{x_i\}_{i=1}^{N}$: Batch of data samples
    \Statex $p_z$: Latent source distribution
    \Statex $d(\cdot, \cdot)$: Distance function
    \Statex $M$: IMLE sample size

\State $f_{\theta_{\perp}} \leftarrow f_{\theta}$ \Comment{Create stop-gradient model copy}
\Statex
\Statex \Comment{\textbf{Calculate Losses}}
\State $\mathcal{L}_{\text{rec}} \leftarrow \frac{1}{N} \sum_{i=1}^{N} d(x_i, f_{\theta}(x_i))$
\Statex
\State Sample $\{z_j\}_{j=1}^{M} \sim p_z$
\State $\mathcal{L}_{\text{IMLE}} \leftarrow \frac{1}{N} \sum_{i=1}^{N} \min_{j} d(x_i, f_{\theta}(z_j))$ 
\Statex
\State Sample $\{z_k\}_{k=1}^{N} \sim p_z$
\State $\mathcal{L}_{\text{idem}} \leftarrow \frac{1}{N} \sum_{k=1}^{N} d(f_{\theta}(z_k), f_{\theta_{\perp}}(f_{\theta}(z_k)))$

\Statex
\Statex \Comment{\textbf{Update Parameters}}
\State $\mathcal{L}_{\text{total}} \leftarrow \mathcal{L}_{\text{rec}} + \mathcal{L}_{\text{IMLE}} + \mathcal{L}_{\text{idem}}$
\State Update $\theta$ via gradient descent on $\mathcal{L}_{\text{total}}$
\end{algorithmic}
\end{algorithm}

\subsection{Manifold Distance and Density Estimation}\label{sec:mdlfdens}
    Empirically, we observe that NAIGNs implicitly infer the manifold distance. 
    Specifically, the learned drift resembles the manifold distance field, at least locally around the data manifold (see Section \ref{subssection:exp–manifold_distance_learning}). 
    This is consistent with the two design conditions satisfied by NAIGNs, namely, that they behave as the identity function on the data manifold, and map other points onto it. 
    However, the manifold distance is not the unique function satisfying these properties. 
    Therefore, the emergence of the distance field in NAIGNs is a peculiar phenomenon. 
    We attribute this, speculatively, to the tendency of  neural networks to learn smooth functions, either due to explicit regularization or through the implicit biases of gradient-based training. 
    More specifically, it is known in the literature that over parameterized autoencoders not only fix the data, but additionally attract points in the ambient space to the closest datapoint \cite{radhakrishnan2020overparameterized}. Since we deploy an autoencoder-like architecture for $f_\theta$, this provides a concrete explanation the emergence of distance fields in NAIGNs.     
    
\begin{figure*}[t]
    \centering
    

    \begin{subfigure}[t]{0.48\linewidth}
        \centering
         \includegraphics[width=1.\linewidth, trim={0pt 35pt 300pt 0pt }, clip]{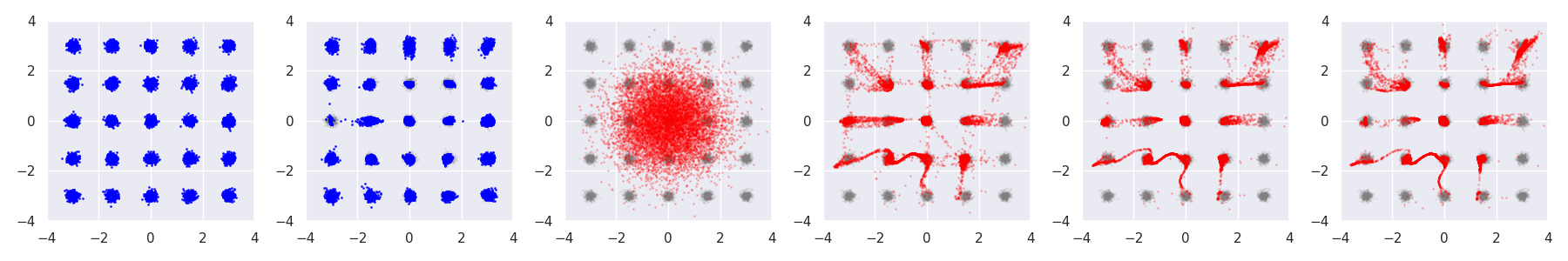}
        \label{fig:sub1}
    \end{subfigure}
    \hfill
    \begin{subfigure}[t]{0.48\linewidth}
        \centering
         \includegraphics[width=1.\linewidth, trim={0pt 35pt 300pt 0pt }, clip]{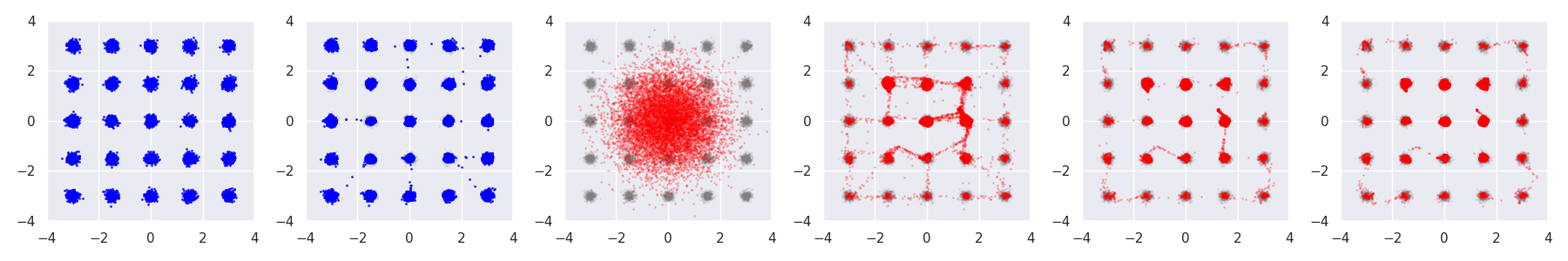}
        \label{fig:sub2}
    \end{subfigure}
    \begin{subfigure}[t]{0.48\linewidth}
        \centering
         \includegraphics[width=1.\linewidth, trim={0pt 35pt 300pt 0pt }, clip]{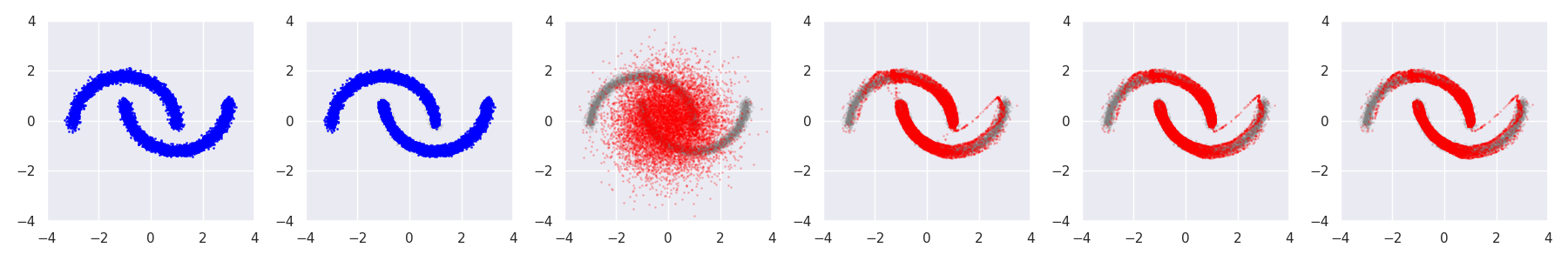}
        \label{fig:sub3}
    \end{subfigure}
    \hfill
    \begin{subfigure}[t]{0.48\linewidth}
        \centering
        \includegraphics[width=1.\linewidth, trim={0pt 35pt 300pt 0pt }, clip]{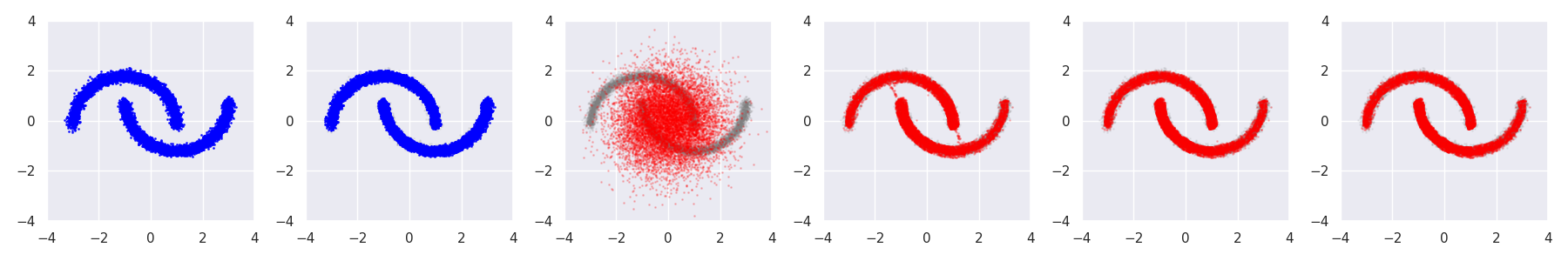}
        \label{fig:sub4}
    \end{subfigure}
    \begin{subfigure}[t]{0.48\linewidth}
        \centering
        \includegraphics[width=1.\linewidth, trim={0pt 35pt 300pt 0pt}, clip]{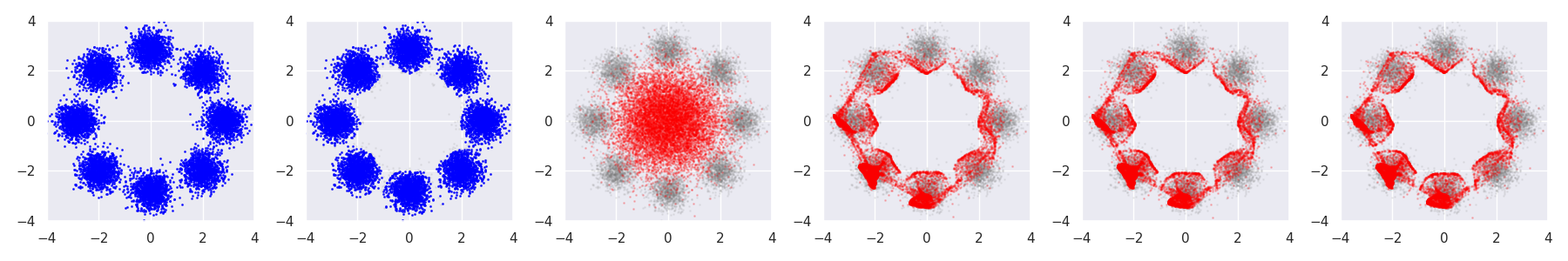}
        \label{fig:sub3}
    \end{subfigure}
    \hfill
    \begin{subfigure}[t]{0.48\linewidth}
        \centering
      \includegraphics[width=1.\linewidth, trim={0pt 35pt 300pt 0pt }, clip]{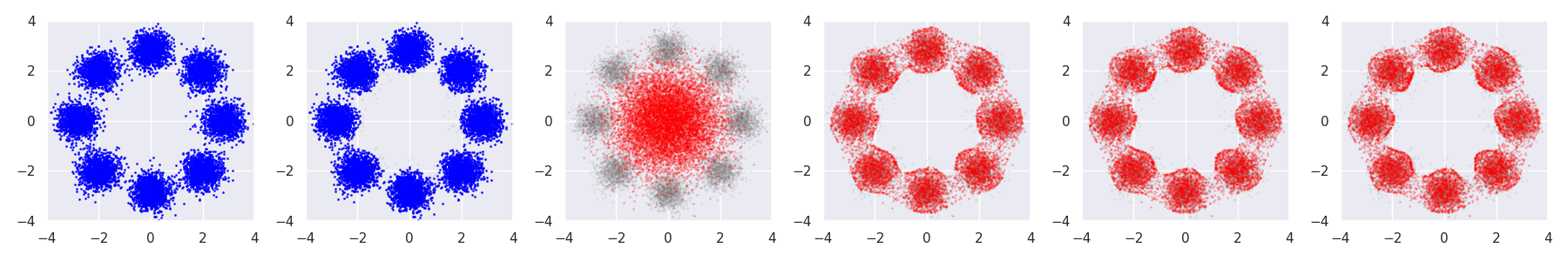}
        \label{fig:sub4}
    \end{subfigure}
 
    \begin{picture}(0,0)
        \put(-237,165){\scriptsize{$p_{\text{data}}(\mathbf{x})$}}
        \put(-187,165){\scriptsize{$f_\theta(\mathbf{\mathbf{x}})$}}
        \put(-140,165){\scriptsize{$p_{Z}(\mathbf{z})$}}
        \put(-90,165){\scriptsize{$f_\theta(\mathbf{z})$}}
        \put(-50,165){\scriptsize{$f_\theta(f_\theta(\mathbf{z}))$}}
        
        \put(26,165){\scriptsize{$p_{\text{data}}(\mathbf{x})$}}
        \put(76,165){\scriptsize{$f_\theta(\mathbf{\mathbf{x}})$}}
        \put(125, 165){\scriptsize{$p_{Z}(\mathbf{z})$}}
        \put(175, 165){\scriptsize{$f_\theta(\mathbf{z})$}}
        \put(216, 165){\scriptsize{$f_\theta(f_\theta(\mathbf{z}))$}}
    \end{picture}
     
    \caption{Mode coverage for IGN (left) and NAIGN (right).}
    \label{fig:mode_issues_ign_vs_naign}

    \vspace{1em}
\end{figure*}
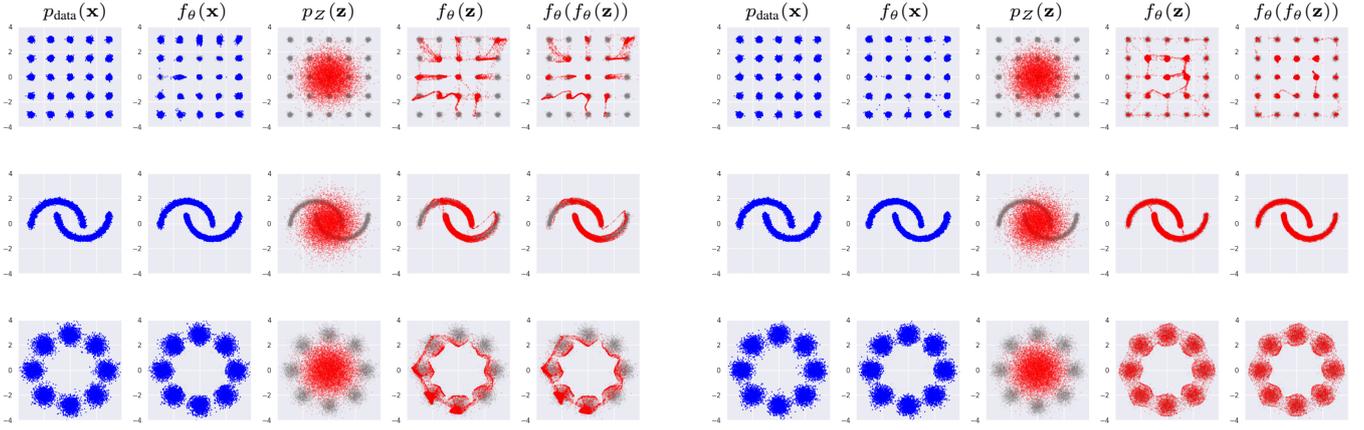

    The fact that the drift measures the proximity to the data manifold can be exploited to design unnormalized density estimators by repurposing NAIGNs as Energy-Based Models (EBMs) \citep{lecun_ebms}. The latter estimate the density of the data distribution $p_{\text{data}}$ as: 
    \begin{equation}
            \hat{p}_\theta(\mathbf{z}) = \frac{e^{-E_\theta(\mathbf{z})}}{Z_\theta},
        \end{equation}
        where \( Z_\theta = \int \exp(-E_\theta(\mathbf{z})) \, d\mathbf{z} \) is the normalizing constant, ensuring a valid probability distribution. The goal is to learn an energy function \( E_\theta(\mathbf{z}) \) that assigns low energy to points near the data manifold \( \mathcal{M} \) and high energy to points farther away, concentrating probability mass on valid data points.
        Given that manifold distance fields exhibit the same functional behavior, we propose transforming the learned drift -- which resembles \( d_{\mathcal{M}} \) -- into an energy function:
        
        \begin{equation}
            E_\theta(\mathbf{z}) = g(\delta_\theta(\mathbf{z})).
        \end{equation}
        
        Here, \( g \colon \mathbb{R}_{\geq 0} \rightarrow \mathbb{R} \) is a rapidly-increasing function such that $g(0) = 0$. This ensures the energy remains low in the proximity of the manifold, and increases swiftly as \( \mathbf{z} \) moves away, aligning with the desired behaviour in EBMs. We propose the following form for $g$:
        \begin{equation}\label{eq:densitypar}
                 g(t) = e^{kt} - 1,
        \end{equation}
        where \( k > 0 \) controls the rate of growth. Since \( E_\theta(\mathbf{z}) = 0 \) for $\mathbf{z} \in \mathcal{M}$, the unnormalized density \( \hat{p}_\theta(\mathbf{z}) \propto e^{-E_\theta(\mathbf{z})}\) equals 1 for data points on the manifold and decreases super-exponentially as \( \mathbf{z} \) moves away. This motivates the interpretation of \( \hat{p}_\theta(\mathbf{z}) \) as a rapidly-decreasing measure of proximity to the data manifold. Note that the normalizing constant \( Z_\theta \) is intractable to compute. Still, the model can generate samples directly through the learned mapping \( f_\theta(\mathbf{z}) \), where \( \mathbf{z} \sim p_Z\). The unnormalized density can be used for tasks where relative likelihood comparisons are sufficient, such as out-of-distribution detection. 

\section{Experiments}\label{sec:exps}
In this section, we provide an empirical investigation of NAIGN, comparing it to IGN on synthetic and real-world data. As an ablation, we also include NAIGN without the idempotent loss (see Equation \ref{eq:ourloss}), referred to as NAIGN$^-$. 

\paragraph{Datasets.} We use both 2D synthetic datasets and real-world image datasets, including MNIST \citep{mnist} and FFHQ\textendash100, a 100-image subset of FFHQ \citep{karras2019style} commonly used to benchmark few-shot generative modelling in low-data regimes \citep{aghabozorgi2023adaptive, rs-imle}. For 2D datasets, we use \textit{2moons} and \textit{8gaussians} \citep{grathwohl2018ffjord}, and we also introduce \textit{grids}, a mixture of 25 Gaussians arranged on a $5\times5$ grid. For high-dimensional cases, MNIST and FFHQ\textendash100 are resized to $32\times32$ and $64\times64$, respectively.


\paragraph{Model Architecture and Training Details.}
For the low-dimensional datasets, we use an MLP with three hidden layers of 512 units. For MNIST and FFHQ-100, we adopt an architecture similar to the original IGN implementation \citep{shocher2023idempotent}. Models are trained with AdamW (learning rate $1\text{e}{-4}$) for 100{,}000 epochs on low-dimensional datasets and 10{,}000 epochs on FFHQ-100. We use batch sizes of 512 (low-dimensional), 256 (MNIST), and 10 (FFHQ-100). The distance metric $d$ in Eq.~\ref{eq:ourloss} is $\ell_2$ for low-dimensional settings and $\ell_1$ for images to encourage sharper reconstructions. For NAIGN on FFHQ-100, we warm-start for 5{,}000 iterations using only the reconstruction loss. We employ a replay buffer (sampling probability $0.5$) that stores mini-batch pairs $(\mathbf{x}, \mathbf{z}^\ast)$, where $\mathbf{z}^\ast=\arg\min_{\mathbf{z}\in\{\mathbf{z}_1,\dots,\mathbf{z}_M\}} d\!\left(\mathbf{x}, f(\mathbf{z})\right)$ is selected per sample via nearest-neighbor search among $M$ generated candidates. The IMLE hyperparameter is set to $M=10\times$ batch size for low-dimensional and MNIST experiments, and $M=100\times$ batch size for FFHQ-100 (i.e., $M=1000$). Similar to \citep{shocher2023idempotent}, we use a structured prior based on each batch’s Fourier-domain statistics; we also evaluated a standard normal prior and observed similar results. For IGN baselines, we set $w_{\text{rec}}=1$, $w_{\text{idem}}=1$, and $w_{\text{tight}}=0.1$ on low-dimensional datasets, and $w_{\text{idem}}=0.125$ for high-dimensional experiments. We run three seeds on low-dimensional datasets and six seeds on MNIST. All experiments were performed on an NVIDIA DGX A100 system (80\,GB VRAM per GPU).

\paragraph{Evaluation Metrics.} In our experiments, we use the following metrics. The reconstruction is measured by the Mean Absolute Error (MAE). To evaluate generation, we deploy two metrics: Coverage and Density by \citep{prdc} --  which improve on a conventional metric for evaluation of generative models \citep{ipir} -- and a modified version of the Fréchet Inception Distance \citep{heusel2017gans}) where the Fréchet distance is computed between real and the generated images in the latent space of a pre-trained LeNet model. Hereafter, the modified Fréchet Inception Distance is referred to as FLD.

\subsection{Addressing Mode and Instability Issues of IGN}  

We begin by highlighting the mode issues and training instabilities that IGNs are susceptible to, and we demonstrate how our proposed method, NAIGN, overcomes these problems. 

Figure \ref{fig:mode_issues_ign_vs_naign} (left) displays evidence of mode issues of IGN. The top-left row (\textit{grids}) exhibits mode dropping since the outermost modes of the dataset are not covered on both the first and the second model iterations. 
Similarly, in the middle-left row (\textit{2moons}), we observe that IGN suffers from partial mode coverage. 
The bottom-left row (\textit{8gaussians}) exhibits mode collapse issues, where IGN converges on the closest data points to the source distribution.
Despite these issues, IGN performs well in preserving reconstructed data and preserving data manifold. In contrast, our method, NAIGN, avoids these problems, as demonstrated by the right-side rows of the same figure. 

\begin{figure}[ht]
    \centering
    \includegraphics[width=.95\linewidth]{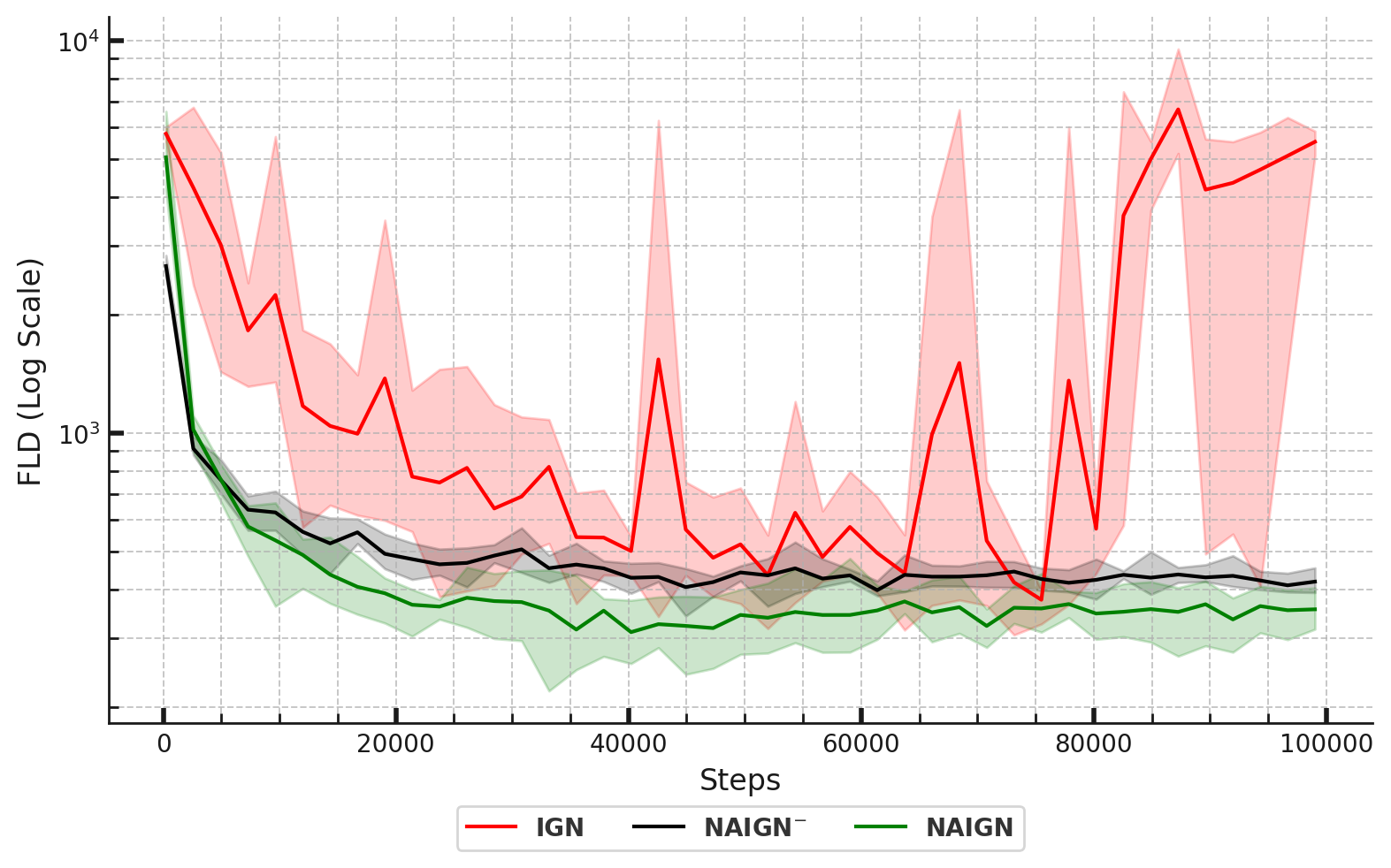}
    \caption{FLD scores for methods during training. Shaded region contains min and max.}
    \label{fig:mmdfidplot}
    \vspace{2em}
\end{figure}

In Figure \ref{fig:mmdfidplot}, we report the FLD score between a generated set of samples and the test dataset of MNIST during training. The plot illustrates the training instabilities of IGN. The training curve initially decreases steadily but stabilizes around 40,000 iterations, after which it diverges drastically. Instead, the curve for NAIGN decreases smoothly, demonstrating the training stability of our proposed method.  

\subsection{Performance Evaluation}
    In our next set of experiments, we evaluate the generation and restoration performance of NAIGN, NAIGN$^-$, and IGN on the MNIST dataset. We consider the following evaluation scenarios.  
    
    \begin{itemize}[wide, labelindent=0pt]
    
    \item \textit{Generation}: We measure the generative performance via Coverage, Density, and FLD between 10000 generated samples $f_\theta(\mathbf{z})$, $z \sim p_Z$, and the first 10000 samples of the test dataset. Additionally, we extend this evaluation to measure the quality of the images produced by the model when applied a second time to the generated samples, i.e., $f_\theta(f_\theta(\mathbf{z}))$, $z \sim p_Z$.


    \begin{table}[ht!]
        \centering
        \vspace{1em}
        \caption{Generation evaluation on MNIST via FLD, Coverage, and Density.}
        \resizebox{\columnwidth}{!}{%
        \begin{tabular}{lccc}
            \toprule
            Method & FLD $(\downarrow)$ & Coverage $(\uparrow)$ & Density $(\uparrow)$ \\
            \midrule
            \midrule
            IGN: $f_{\theta}(\mathbf{z})$ & 359±40 & 0.39±0.04  & 0.38±0.03   \\
            IGN: $f_{\theta}(f_{\theta}(\mathbf{z}))$ & 440±194 & 0.36±0.01  & 0.38±0.04  \\
            \midrule
            NAIGN: $f_{\theta}(\mathbf{z})$ & \textbf{279±40} & 0.58±0.01  & 52±0.02   \\
            NAIGN: $f_{\theta}(f_{\theta}(\mathbf{z}))$ & \textbf{247±34} & 0.56 ± 0.02  & 49±0.02   \\
            \midrule
            NAIGN$^-$: $f_{\theta}(\mathbf{z})$ & 382±32 & \textbf{0.60±0.01} & \textbf{0.63±0.2}  \\
            NAIGN$^-$: $f_{\theta}(f_{\theta}(\mathbf{z}))$ & 351±27 & \textbf{0.60±0.01} & \textbf{0.60±0.02} \\
            \bottomrule
        \end{tabular}%
        }
        \label{table:gen_rec_eval}
    \end{table}

    \item \textit{Restoration}: We measure restoration capabilities via MAE between degraded data points $\mathbf{x}'$ and their restored outputs $f_\theta(\mathbf{x})$. The degradations we consider are Gaussian noise, Gaussian blurring, salt and pepper noise, and random deletion of rows and columns akin to inpainting tasks. 
    \end{itemize}

    \begin{table}[h!]
        \centering
        \caption{Reconstruction evaluation on MNIST via MAE ($\downarrow$).}
        \resizebox{\columnwidth}{!}{%
        \begin{tabular}{lcccc}
                \toprule
                Method & Blur & Gaussian Noise & Salt\&Pepper & LinesRows \\
                \midrule
                \midrule
                IGN         &  0.44±0.43   &  0.31±0.10   & 0.27±0.06   & 0.29±0.29 \\
                \midrule
                NAIGN       &  \textbf{0.15±0.01}   &  \textbf{0.21±0.03}   & \textbf{0.22±0.03}   &  \textbf{0.11±0.00} \\
                \midrule
                NAIGN$^-$   &  \textbf{0.15±0.00}   &  0.26±0.02   & 0.25±0.02   & \textbf{0.11±0.00} \\
                \bottomrule
        \end{tabular}%
        }
        \label{table:recon_eval}
    \end{table}

\begin{figure*}[ht!] 
    \centering 
    \begin{subfigure}[b]{0.45\textwidth}
        \centering
        \includegraphics[width=\linewidth]{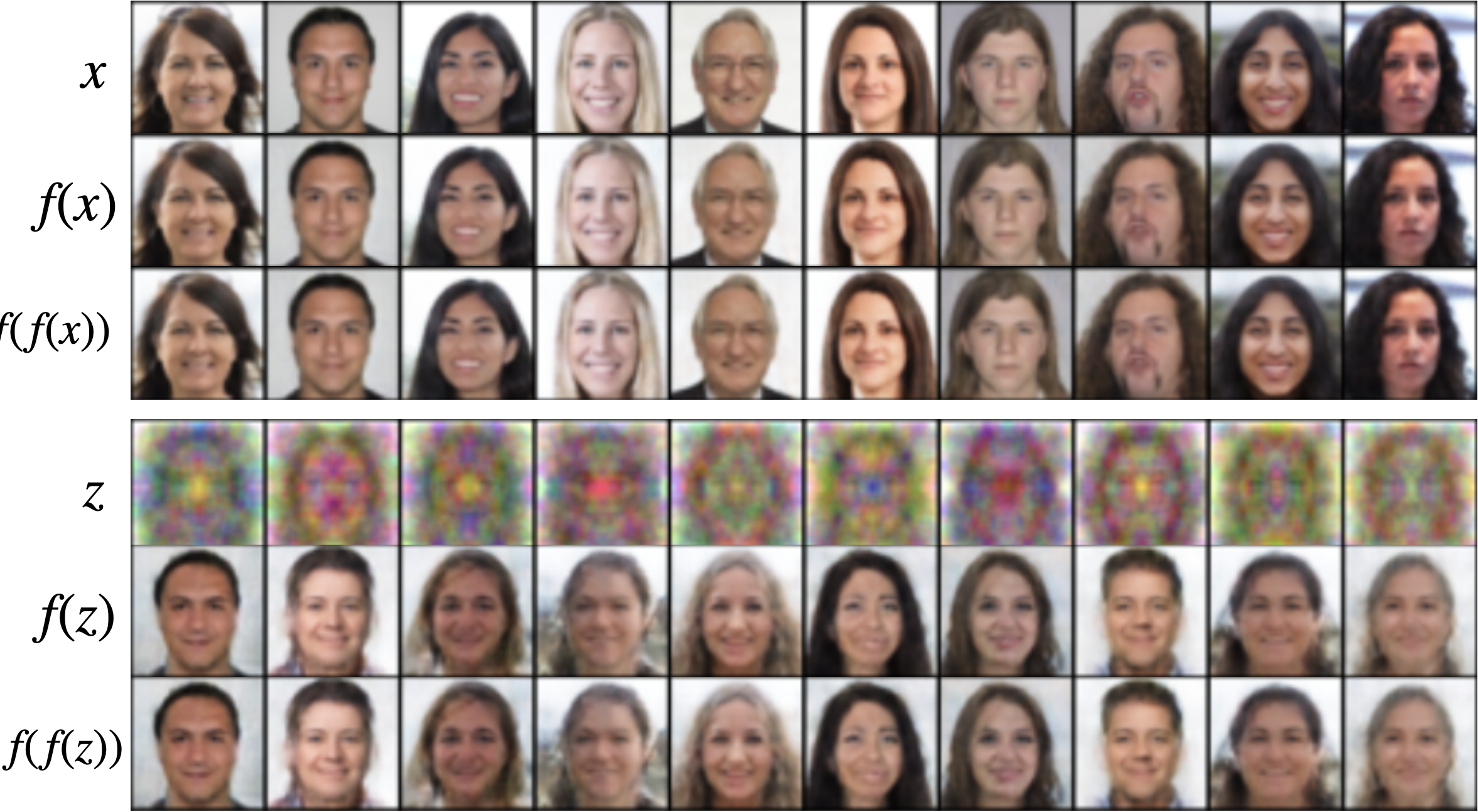} 
        \label{fig:sub1}
    \end{subfigure}
    \hfill 
    \begin{subfigure}[b]{0.45\textwidth}
        \centering
        \includegraphics[width=\linewidth]{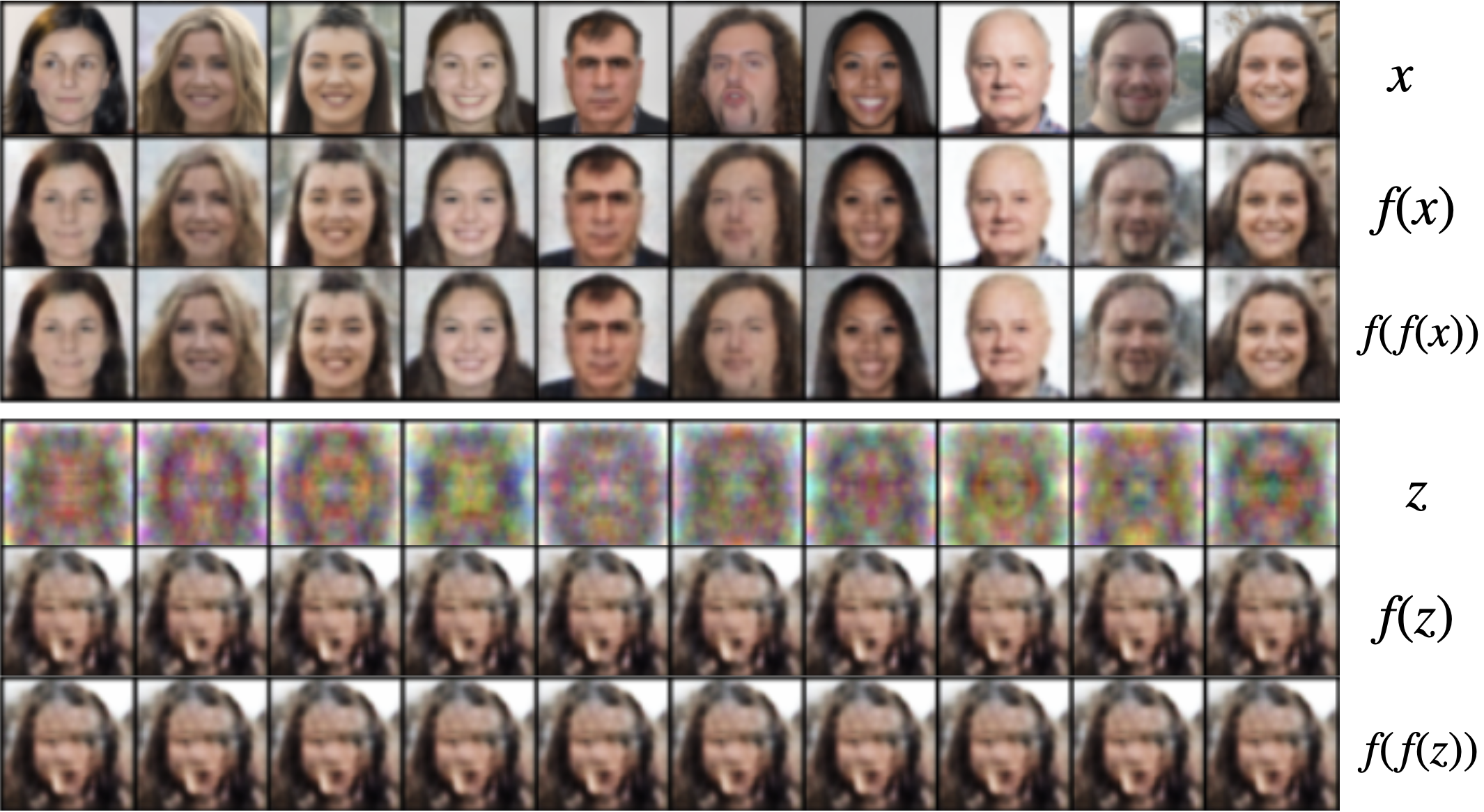} 
        \label{fig:sub2}
    \end{subfigure}
    \caption{{Reconstructed and generated samples from NAIGN (left) and IGN (right), both trained on the FFHQ-100 dataset. While NAIGN demonstrates faithful reconstruction and generative diversity, IGN consistently exhibits mode collapse.}}
    \vspace{1em}
    \label{fig:fewshot_comparison}
\end{figure*}

Table \ref{table:gen_rec_eval} and \ref{table:recon_eval} report generation and restoration results, respectively. We observe that NAIGN outperforms IGN in all the scenarios. NAIGN$^-$ exhibits worse performance than IGN on generation, since the FLD score is significantly lower, while Coverage and Density are comparable. Moreover, NAIGN$^-$ and IGN are comparable on restoration performance. We conclude that NAIGN should be generally preferred over NAIGN$^-$, since it exhibits overall better generative performance, while still projecting points close to the data manifold, such as degraded data, back to it.

Additionally, we provide model output samples for reconstruction, generation, and restoration tasks in Section~4 of the supplementary material, which is available at \cite{naign2025supp}.

\begin{figure}
    \centering
    \begin{subfigure}[b]{0.3\columnwidth}
        \includegraphics[width=\textwidth]{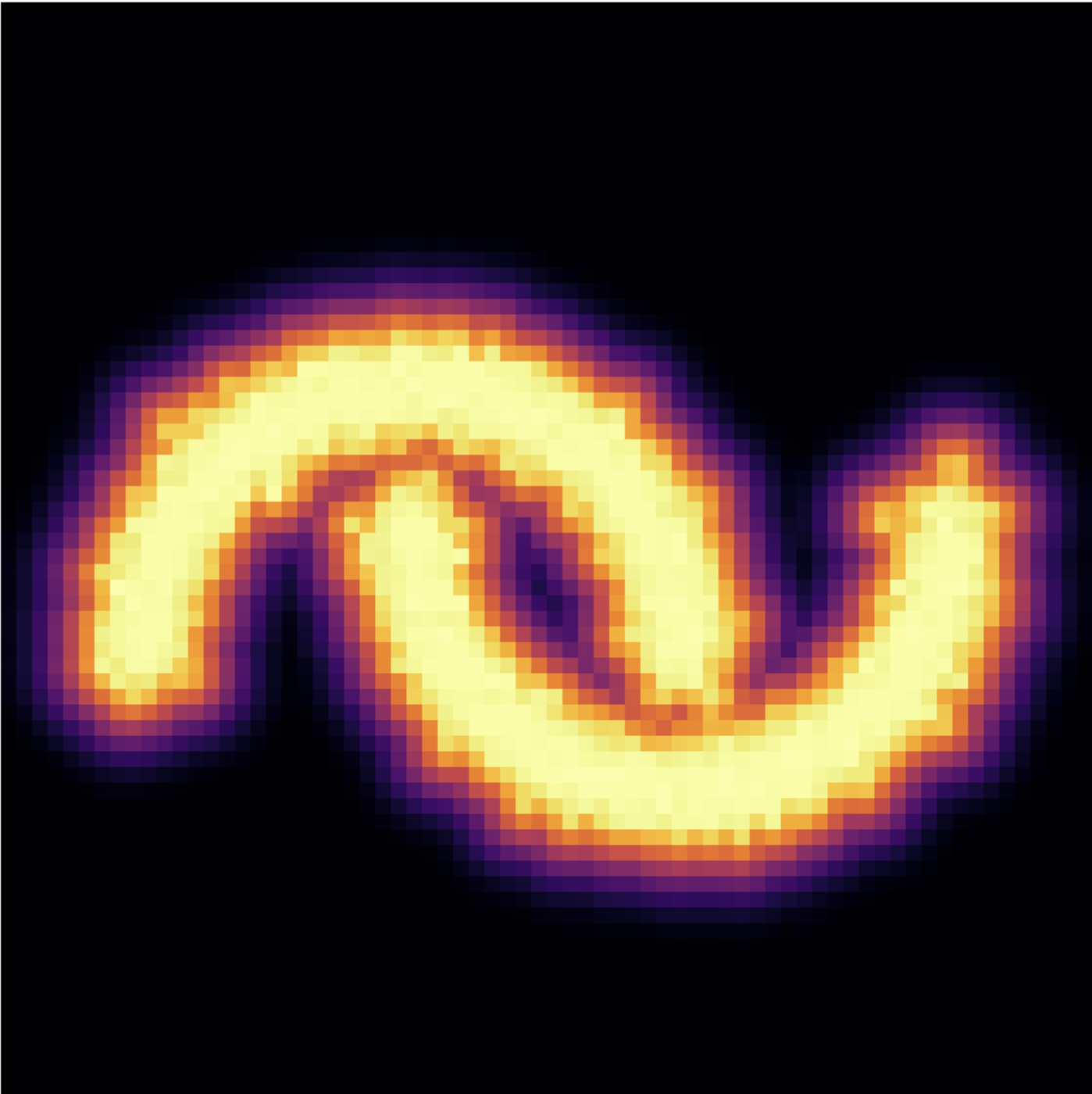}
    \end{subfigure}
    \hfill
    \begin{subfigure}[b]{0.3\columnwidth}
        \includegraphics[width=\textwidth]{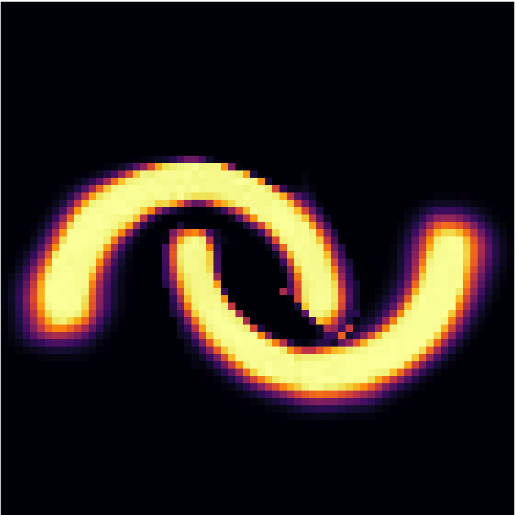}
    \end{subfigure}
    \hfill
    \begin{subfigure}[b]{0.3\columnwidth}
        \includegraphics[width=\textwidth]{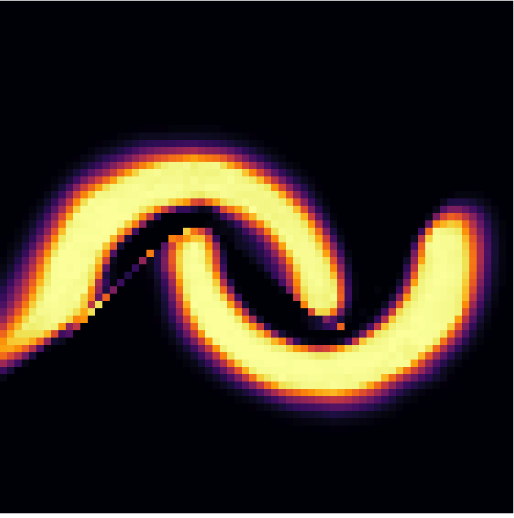}
    \end{subfigure}
    \vspace{1em}
    \caption{Unnormalized density estimation using NAIGN (middle) and IGN (right), compared to the unnormalized density generated from the true manifold distance field (left).}
    \label{fig:densityest}
\end{figure}

\subsection{Few-Shot Image Synthesis}
To further analyze the stability, generalizability and data-efficiency of our proposed model, we conduct an experiment on FFHQ-100, a realistic benchmark with extreme data scarcity \citep{aghabozorgi2023adaptive, rs-imle}. NAIGN demonstrated robust performance, successfully converging with seemingly low amounts of memorization. Instead, IGN, despite an extensive hyperparameter search focusing on the relative loss weights, consistently exhibited training instability. It either suffered from complete mode collapse, resulting in poor generative coverage, or unstably cycled through a small number of modes. 
Figure \ref{fig:fewshot_comparison} displays images generated by both the models, where the performance of the two models can be evaluated qualitatively. Given the small dataset size ($N=100$), quantitative metrics such as FID are not applicable.

\subsection{Manifold Distance Learning}
\begin{figure}
    \centering
    \begin{subfigure}[b]{0.49\columnwidth}
        \includegraphics[width=\textwidth]{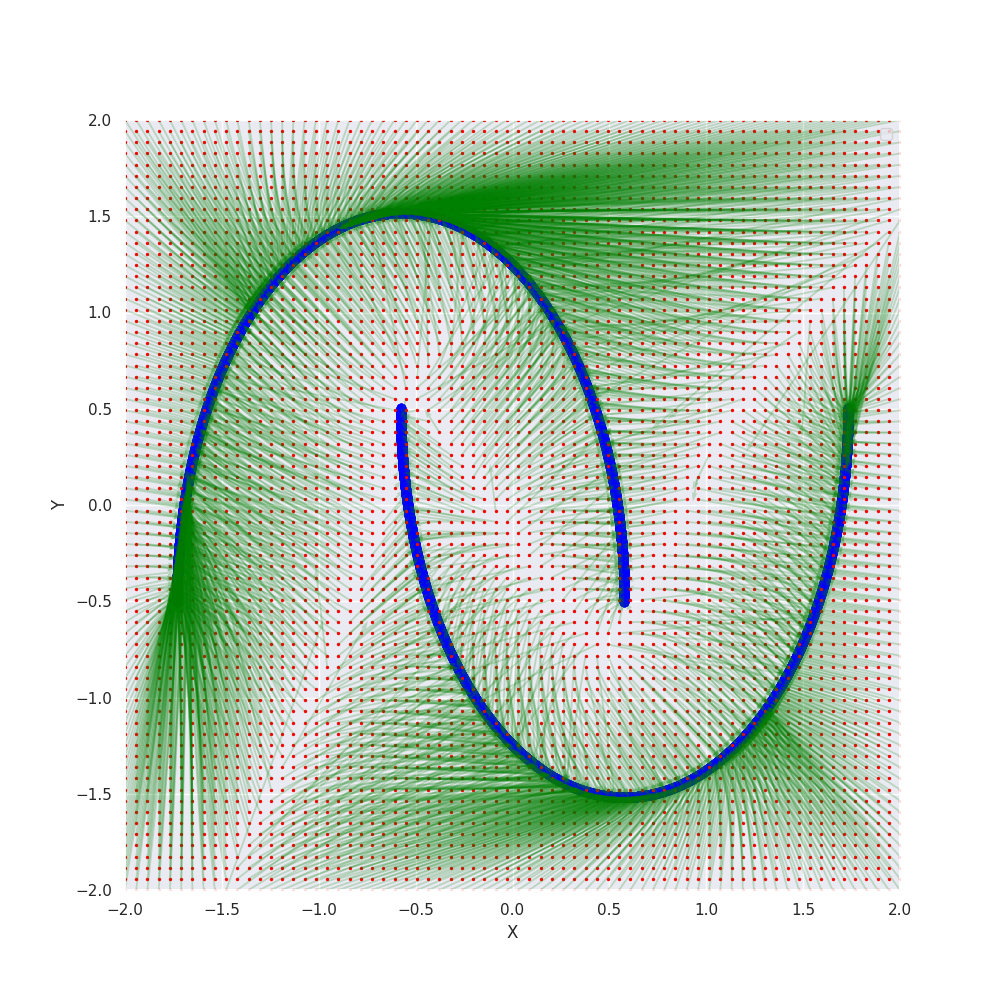}
    \end{subfigure}
    \hfill
    \begin{subfigure}[b]{0.49\columnwidth}
        \includegraphics[width=\textwidth, trim={0 0 0 100pt}, clip]{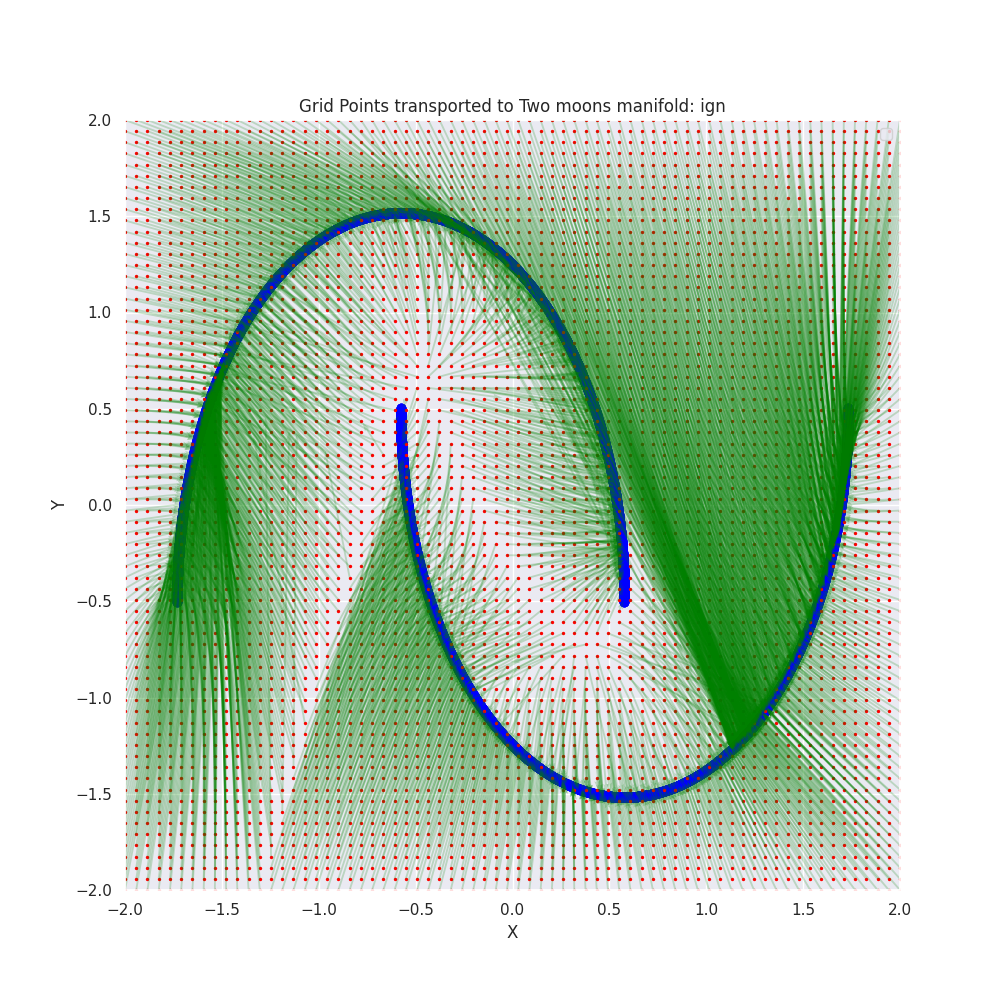}
    \end{subfigure}
    \caption{Projection maps of NAIGN (left), and IGN (right) trained on the \textit{2moons} dataset with zero noise (blue). Red dots are the inputs into the network and the green arrows indicate where these get projected onto.}
    \label{fig:projectionmaps}
    \vspace{2em}
\end{figure}

{\label{subssection:exp–manifold_distance_learning}}
In this section, we demonstrate that NAIGN implicitly learns an approximation to the manifold distance field. Concretely, the learned drift \( \delta_\theta(\mathbf{z}) = d(\mathbf{z}, f_\theta(\mathbf{z})) \) approximates \( d_\mathcal{M}(\mathbf{z}) \) (see Equation~\ref{eq:distfield}) for all \( \mathbf{z} \in \mathbb{R}^N \). We illustrate this in Figure~\ref{fig:projectionmaps}, which shows snapshots of the drift landscape learned by NAIGN during training on the \textit{2moons} dataset. In Figure~\ref{fig:projectionmaps}, we show the \textit{projection map}, where we display the projection of a dense grid of points around and on top of the \textit{2moons} dataset, using both a trained NAIGN model and a trained IGN model. Here, the projection displacement is approximately zero (i.e., fixed-point projection) for grid points lying on the data manifold, and points close to the manifold are roughly projected to their nearest location on it. However, when zooming out, grid points farther from the manifold are sometimes projected back to a nearby point on the manifold. This behavior is less consistent in IGNs, for which the displacement of some distant grid points is mapped far away.

Moreover, Figure \ref{fig:densityest} reports heatmaps of the estimated density at the end of the training run via the procedure described in Section \ref{sec:mdlfdens} ($k=2$ in Equation \ref{eq:densitypar}), via both IGN and NAIGN. In this case, to highlight the density estimation capabilities in a more realistic setting, we deploy a `fuzzy' version of the \textit{2moons} dataset, where Gaussian noise is injected in the data manifold.  
As we can see, both NAIGN and IGN infer distance fields that are close to the ground-truth manifold distance field. 
However,  the density estimated from IGN suffers from artifacts, as evident in the visualization. These artifacts are probably due to IGN's deficiencies in representing and covering faithfully the modes of the distribution, as discussed in Section \ref{sec:idemgen}.

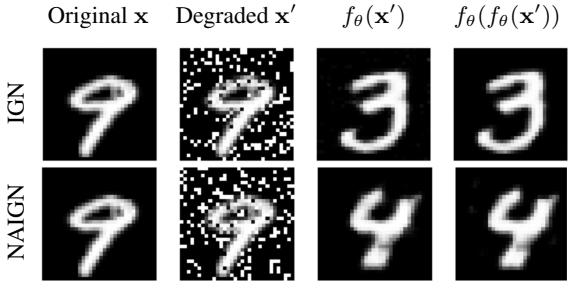
\begin{figure}[t!]
    \centering
    \vspace{2em}
    \includegraphics[width=0.8\linewidth]{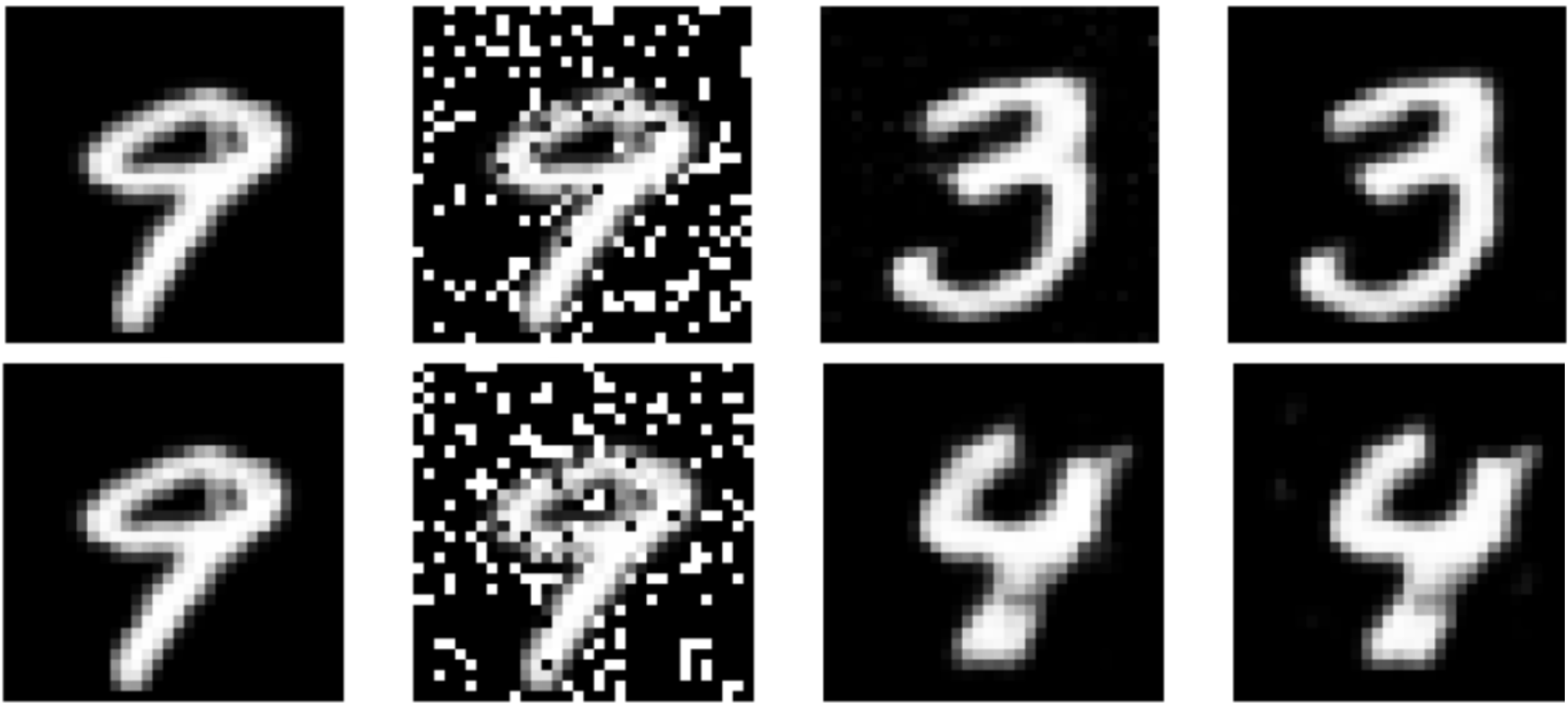}    
    \vspace{1em}
    \caption{Example of manifold projection failure. IGN (top row) and NAIGN (bottom row).}
    \begin{picture}(0,0)
        \put(-95,128){\small{Original $\mathbf{x}$}}
        \put(-47,128){\small{Degraded $\mathbf{x}'$}}
        \put(15,128){\small{$f_\theta(\mathbf{x}')$}}        
        \put(57,128){\small{$ f_\theta(f_\theta(\mathbf{x}'))$}}
        \put(-110, 95){\rotatebox[origin=c]{90}{\small{IGN}}}
        \put(-110, 50){\rotatebox[origin=c]{90}{\small{NAIGN}}}
    \end{picture}
    \vspace{1em}
    \label{fig:shoot}
\end{figure}

Lastly, Figure \ref{fig:shoot} reports an example of a manifold projection failure. In this case, a degraded data point is projected to a point on $\mathcal{M}$ different from the original one. Since the digit `9' becomes, respectively, `3' and `4' for IGN and NAIGN, the failure is less drastic for the latter model. This projection failure has a low-dimensional counterpart near the central tips of both moons in figure \ref{fig:projectionmaps}, where a small perturbation from the tips leads the model to projection to far away points on the other half-moon. 

\subsection{Discussion on the Hyperparameter $M$}\label{sec:hyperimle}
The most expensive component of NAIGN's training procedure is the computation of IMLE, which, for each data sample in a batch of $N$ data, involves a minimum search over $M$ samples. This leads to a complexity of $\mathcal{O}(NM)$. Therefore, the choice of the hyperparameter $M$ is crucial. Here, we investigate the effect of $M$ on the quality of generation, analyzing the trade-off between efficiency and performance. {Conceptually, the number of IMLE samples $M$ must be large enough to effectively cover the modes of the data distribution during training. The precise theoretical scaling of $M$ with respect to a formal measure of data complexity is, to our knowledge, an open research question. The original IMLE paper \citep{imle} establishes that the loss is an unbiased estimator of the log-likelihood, but deriving sample complexity bounds for this process is challenging and goes beyond the scope of the current work.} 

\begin{figure}[h]
    \centering
    \includegraphics[width=.95\linewidth, trim={0pt 0pt 0pt 85pt}, clip]{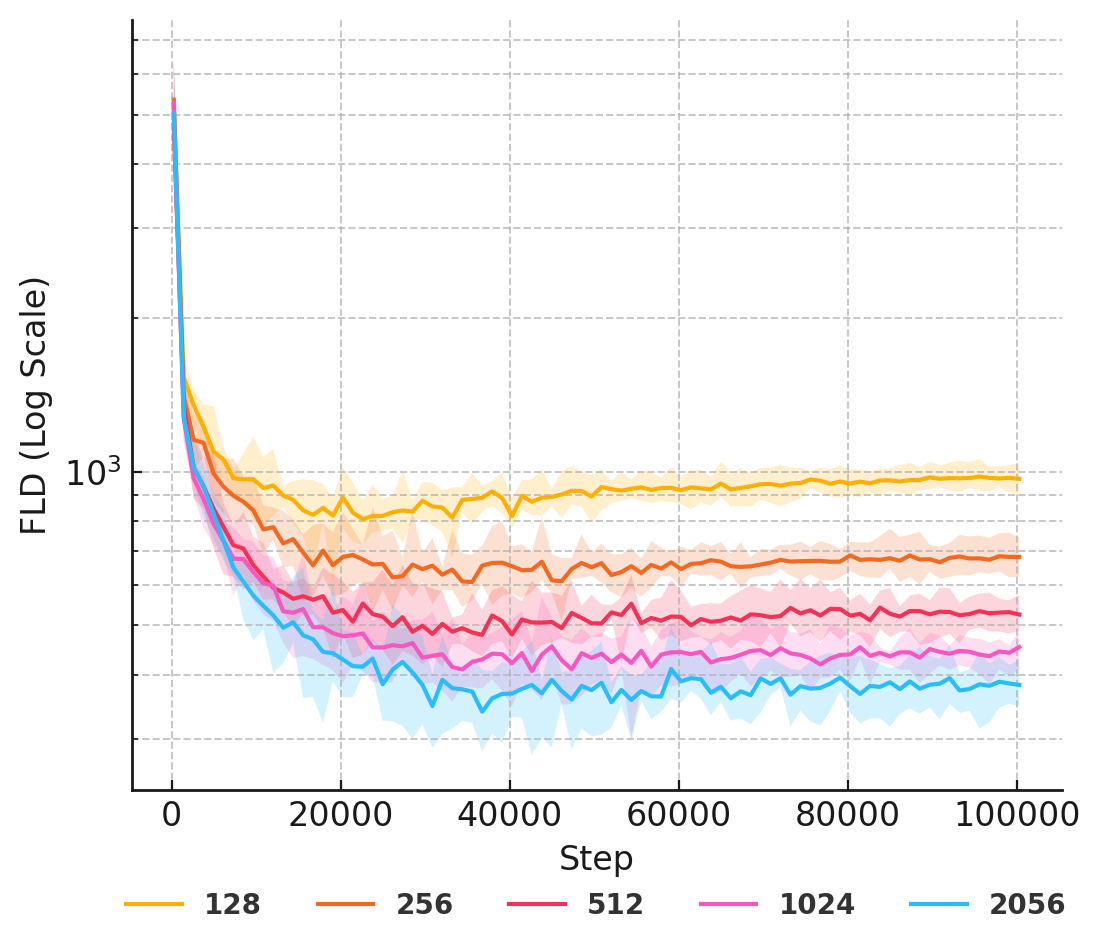}
    \caption{Effect of IMLE hyperparameter $M$ on generative performance throughout training, measured via FLD score.}
    \label{fig:imle_m}
\end{figure}

Instead, in what follows we provide an empirical investigation of the generative performance with respect to carrying $M$ on MNIST. Figure \ref{fig:imle_m} reports the FLD score throughout training for different choices of $M \in \{128, 256, 512, 1024, 2056\}$, with a batch size of $N = 256$. It is evident that an increased $M$ leads to a better performance score, as expected. Moreover, by comparing with Table \ref{table:gen_rec_eval}, $M$ needs to be larger than a certain threshold in order for NAIGN to perform comparably or better to IGN.  

\section{Limitations and Conclusion} 
We introduced Non-Adversarial Generative Idempotent Networks (NAIGNs), offering a novel approach to generative modeling designed around idempotency without adversarial training. While promising, our method has limitations that suggest directions for future work.

\textbf{Loss Function Limitations.} Our reliance on loss functions in raw data space (e.g., pixel differences) may be inefficient for high-resolution data and may not capture semantic similarities effectively. Exploring alternative loss functions or representations that align better with perceptual differences, such as operating in latent space using autoencoders or VAEs, as seen in methods such as latent diffusion models \citep{rombach2022high}, could enhance efficiency and performance.

\textbf{Computational Complexity of IMLE.} NAIGN relies on IMLE, which comes with its own limitations. First, computing the $argmin$ for each batch element per iteration is computationally intensive. This has been addressed by more sophisticated IMLE variants (e.g., rejection sampling IMLE \citep{rs-imle} and Adaptive IMLE \citep{aghabozorgi2023adaptive}), which improve computational efficiency. These variants can be included in NAIGN, potentially reducing the total number of iterations required within the same computational budget. Second, IMLE serves as the generative loss component. Exploring alternative generative loss functions, such as MMD or other suitable distance measures, represents another direction for future investigation. 

\textbf{Connection to Implicit Neural Representations.} Our method parallels Implicit Neural Representations (INR; \citep{de2023deep}) that learn continuous shapes in low dimensions. In traditional INRs, the input is a coordinate point, and the output is either the corresponding value or zero at data positions, learning the surface of an object through its zero level sets. Similarly, our method takes an input vector, in our case, the entire image, and aims for the model to produce minimal drift between the input and output, meaning the distance between them is minimized. This minimal drift corresponds to the zero level set in INRs. Thus, we are implicitly learning the surface of the data manifold, where the zero level set represents the manifold itself. We plan on investigating the suitability of idempotent generative models in learning and generating 3D shapes. 

\textbf{Positioning of Idempotent Generative Networks.}
IGNs represent a promising model class that integrates generative and restorative capabilities. However, we believe their limited application so far stems from critical challenges such as training instability and mode collapse, which our work addresses.
Compared to diffusion models, NAIGN learns a direct manifold projector that maps inputs onto the data manifold in a single forward pass, offering greater compute and latency efficiency than the inherently multi-step sampling of diffusion models. 
Relative to GANs, NAIGN additionally supports \textit{iterative refinement}, whereby an unsatisfactory output can be improved by one or two re-applications of the model. 
Finally, the learned drift function serves as an unnormalized density surrogate, enabling out-of-distribution detection without a separate model. 
Our primary contribution is addressing the critical shortcomings of the original IGN framework; mode collapse, mode dropping, and training instability by replacing the adversarial objective with a stable, non-adversarial objective that encourages full mode coverage. 
These enhancements are essential for making idempotent models competitive with leading generative methods. 
By doing so, NAIGN represents a significant step towards versatile and efficient models that simultaneously act as both robust manifold projectors and high-fidelity generators.

\section*{Acknowledgments}
{This work was partially supported by the Wallenberg AI, Autonomous Systems and Software Program (WASP) funded by the Knut and Alice Wallenberg Foundation, the Swedish Research Council, and the European Research Council. The computations were enabled by Berzelius resources provided by the Knut and Alice Wallenberg Foundation at the National Supercomputer Centre.}

\bibliography{main.bib}

\clearpage
\onecolumn
\section*{Appendix}
\vspace{10em}
\addcontentsline{toc}{section}{Appendix}
\input{supp.tex}
\twocolumn
\end{document}

%% file: math_commands.tex
\usepackage{amsmath,amsfonts,bm}

\def\eqref#1{equation~\ref{#1}}

\def\1{\bm{1}}

\DeclareMathAlphabet{\mathsfit}{\encodingdefault}{\sfdefault}{m}{sl}
\SetMathAlphabet{\mathsfit}{bold}{\encodingdefault}{\sfdefault}{bx}{n}

\newcommand{\Ls}{\mathcal{L}}



%% file: supp.tex






\section{NAIGN Torch Pseudocode}
\label{appendix:torch}

\label{appendix:torch}

\begin{figure}
    \centering
    \includegraphics[width=\linewidth]{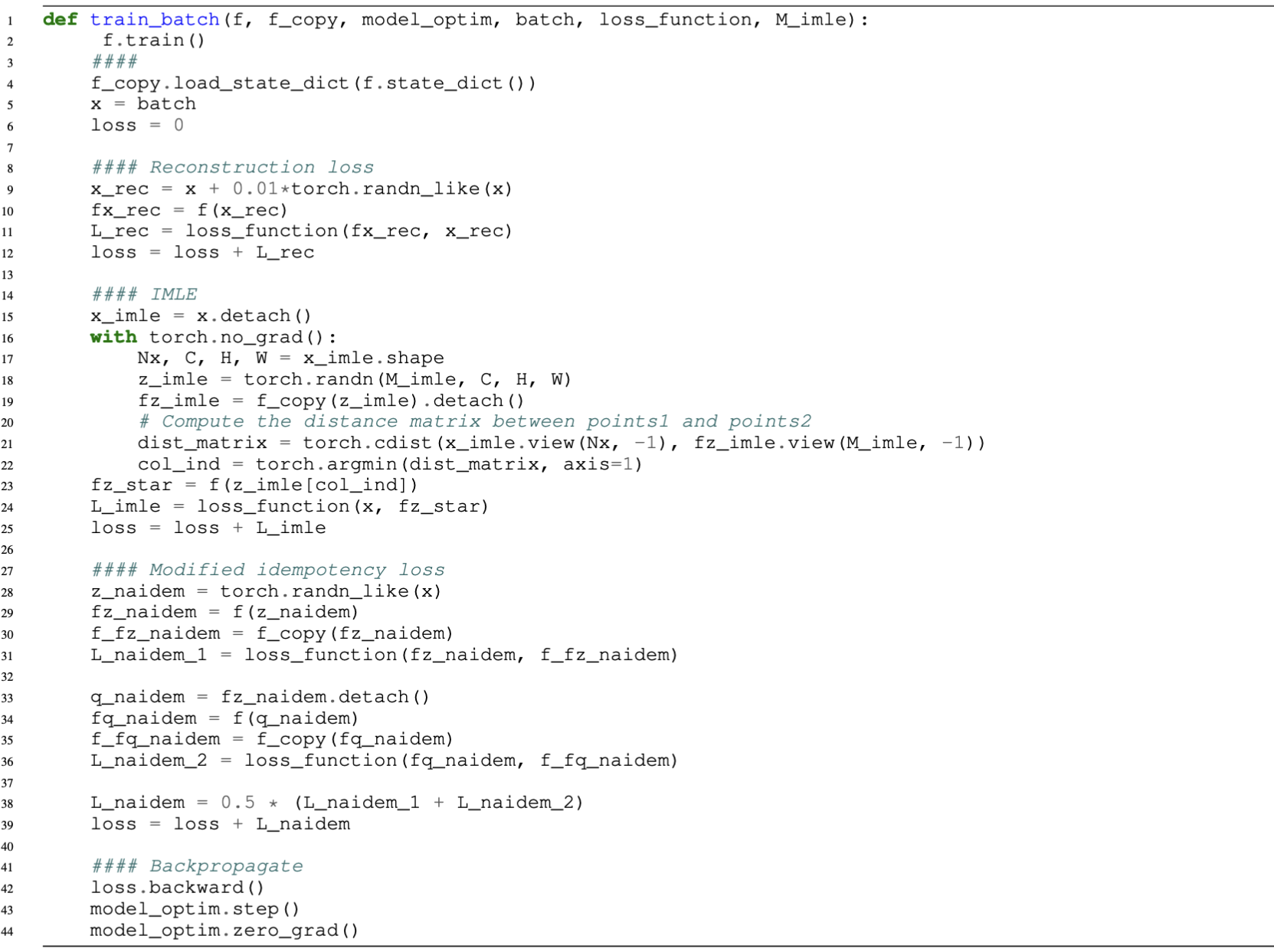}
\end{figure}
    



    
    

\newpage
\section{Modified Idempotent Loss}
\label{appendix:naidem}
The modified idempotent loss in NAIGNs ensures that the model behaves idempotently on data manifold points, source distribution points, and off-manifold points.

The original idempotent loss, introduced by the original IGN paper, enforces idempotency on the source distribution by penalizing discrepancies between the model’s output and its second application:
\[
\mathcal{L}_{\text{idem}}(\theta) = \mathbb{E}_{\mathbf{z} \sim p_{Z}} \left[ d\left( f_{\theta}(\mathbf{z}), f_{\theta_{\bot}} \left( f_{\theta}(\mathbf{z}) \right) \right) \right]
\]
where $\mathbf{z}$ is a sample from the source distribution $p_Z$.

We extend this loss by also applying it to points generated by the model during training, capturing off-manifold behavior. The modified loss is the average of the original idempotent loss and a similar term for model-generated points:
\[
\mathcal{L}_{\text{idem}}(\theta) = 
\frac{1}{2} \left(
\mathbb{E}_{\mathbf{z} \sim p_{Z}} \left[ d\left( f_{\theta}(\mathbf{z}), f_{\theta_{\bot}} \left( f_{\theta}(\mathbf{z}) \right) \right) \right]
+
\mathbb{E}_{\mathbf{q} \sim p_{\theta}} \left[ d\left( f_{\theta}(\mathbf{q}), f_{\theta_{\bot}} \left( f_{\theta}(\mathbf{q}) \right) \right) \right]
\right)
\]
where $\mathbf{q}$ represents samples generated by the model with inputs from the source distribution. This modification ensures the model is trained on both source and off-manifold points, encouraging it to project generated points back onto the manifold. By applying the loss to both types of points, the model becomes explicitly exposed to off-manifold behavior and learns how to act idempotently on such points as well.

\section{Experimental Details: Degradation and Restoration}

We conducted experiments on the MNIST dataset, resized to 32x32 to fit a smaller DCGAN model architecture. We applied four types of degradation: Gaussian blur, Gaussian noise, salt and pepper noise, and random deletion (setting random rows and columns to zero).

\begin{itemize}
    \item  \textbf{Gaussian blur:} We used the Torch transform's Gaussian blur function. The amount of blur was controlled by the kernel size and a blur level parameter (sigma), where the kernel size was computed based on the blur level, and the sigma value matched the blur level.
 
    \item  \textbf{Gaussian noise:} We used a sigma value of 1.0, implemented via the Torch transform's Gaussian noise functionality.

    \item  \textbf{Random row and column deletions:} Each row and column had a 20\% chance of being deleted, with the pixel values set to -1.

    \item  \textbf{Salt and pepper noise:} 20\% of the pixels were deemed corrupted, and of those, 50\% were set to salt (value 1) and 50\% to pepper (value -1), as the MNIST dataset was normalized to the range of -1 to 1.
\end{itemize}

\newpage
\section{Reconstructed, Generated and Restored Sample Images}\label{sec:recimg}
\begin{figure*}[h!]
    \centering
    \begin{subfigure}[b]{0.22\textwidth}
        \centering
        \includegraphics[height=9cm, trim={0pt 0pt 365pt 0pt}, clip]{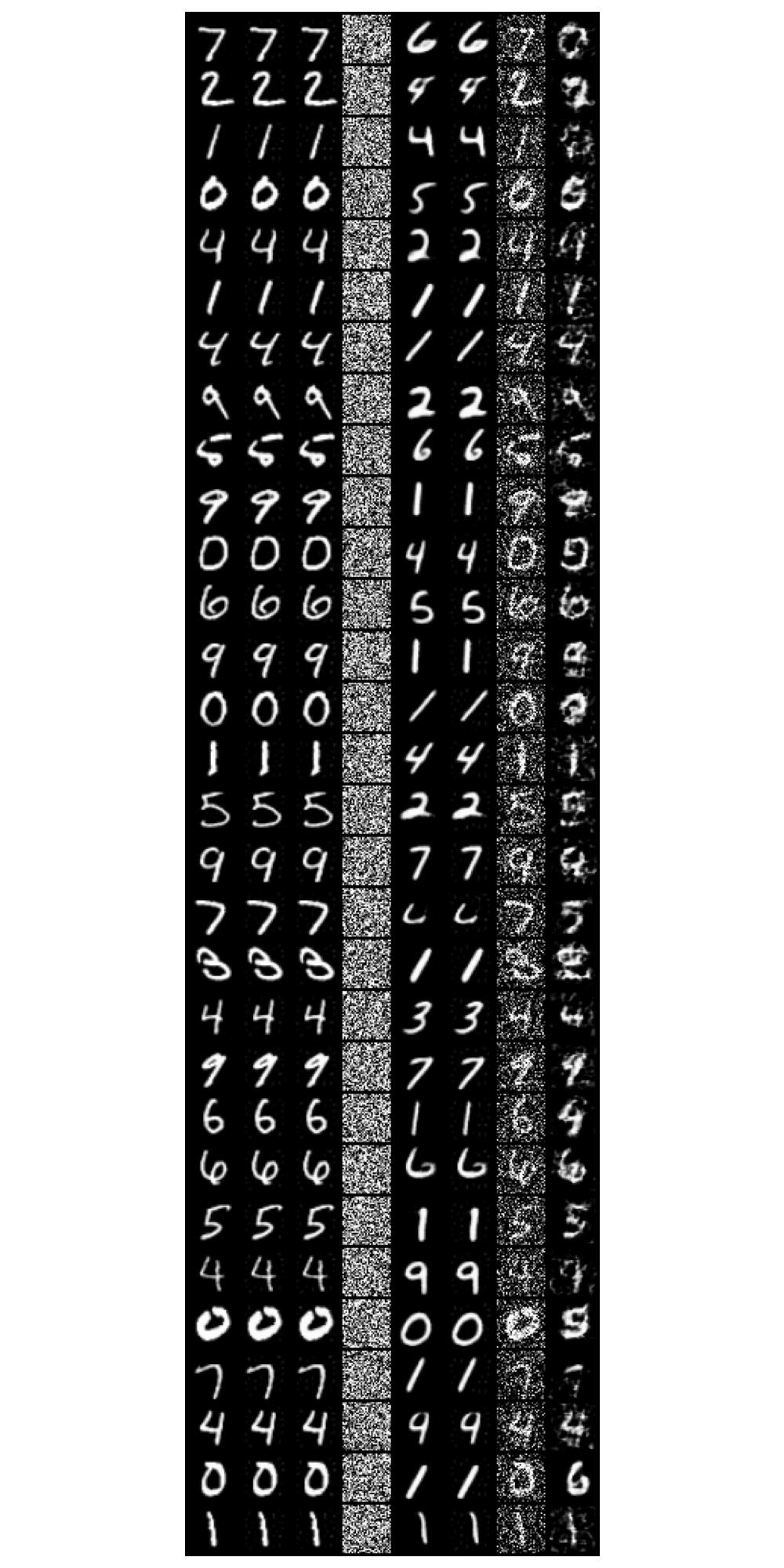}

    \end{subfigure}
    \begin{subfigure}[b]{0.22\textwidth}
        \centering
        \includegraphics[height=9cm, trim={0pt 0pt 365pt 0pt}, clip]{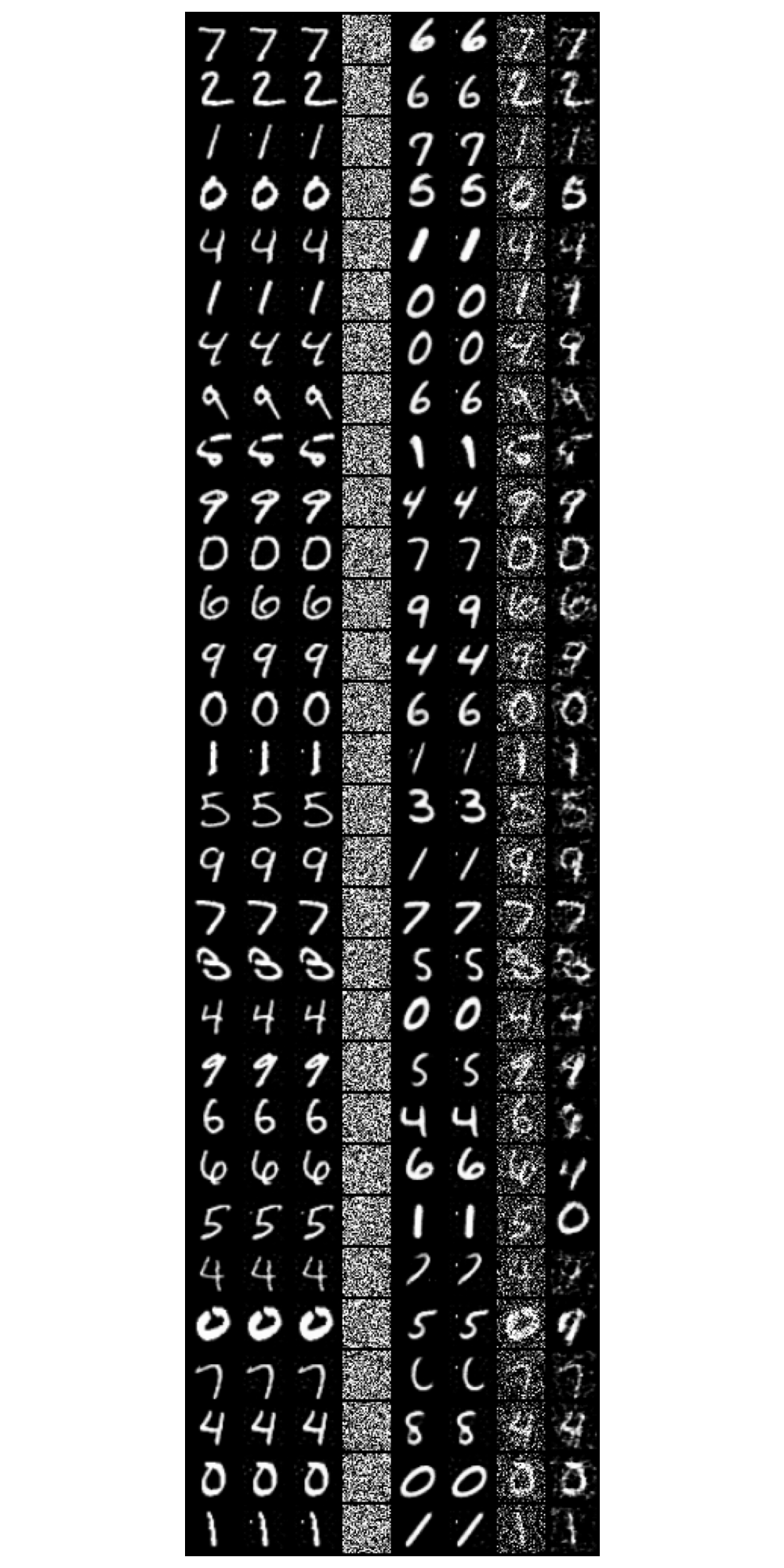}

    \end{subfigure}    
    \begin{subfigure}[b]{0.22\textwidth}
        \centering
        \includegraphics[height=9cm, trim={0pt 0pt 365pt 0pt}, clip]{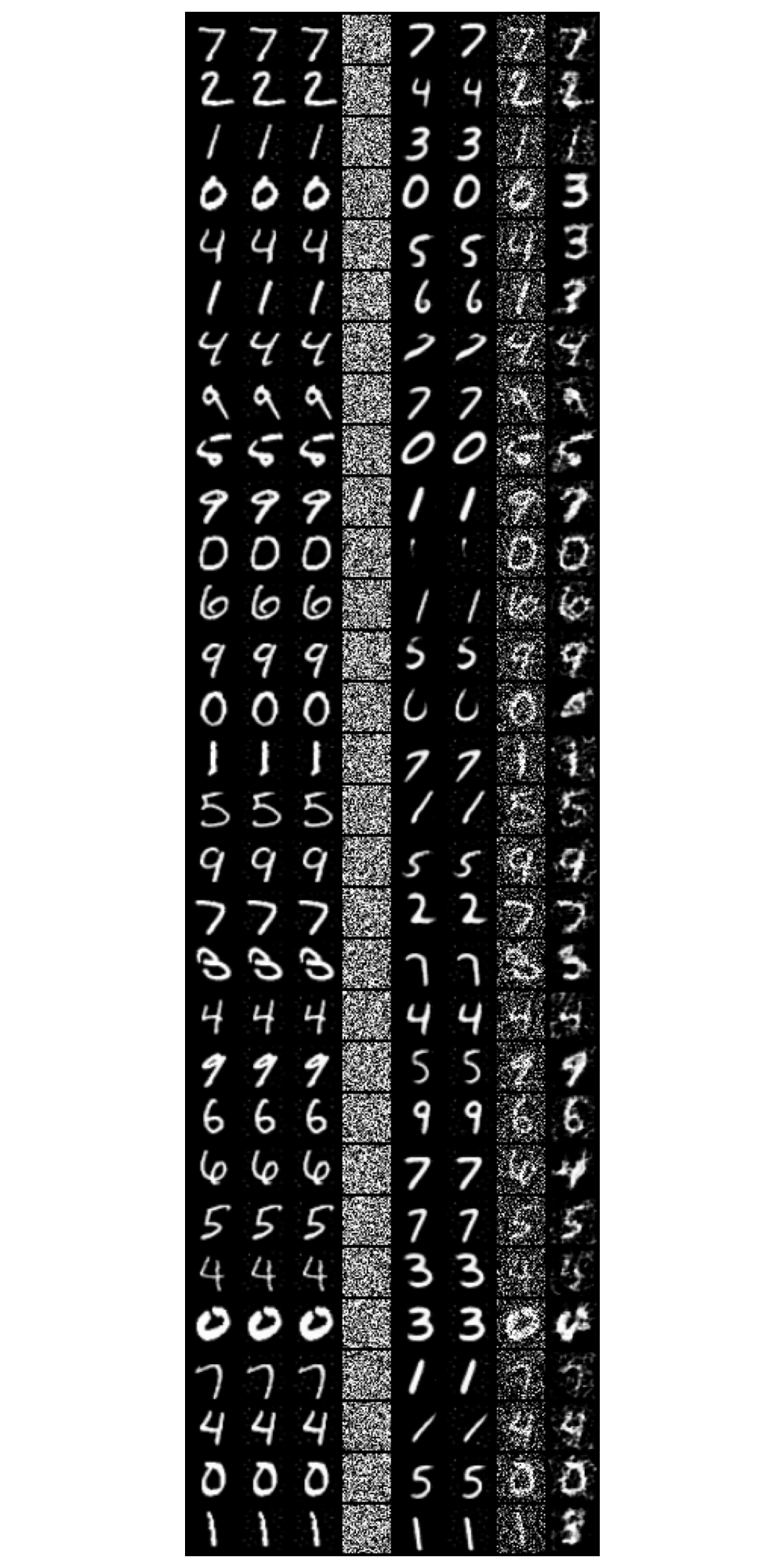}
    \end{subfigure}    
    \hspace{.1em}

    \centering
    \begin{subfigure}[b]{0.22\textwidth}
        \centering
        \includegraphics[height=9cm, trim={0pt 0pt 365pt 0pt}, clip]{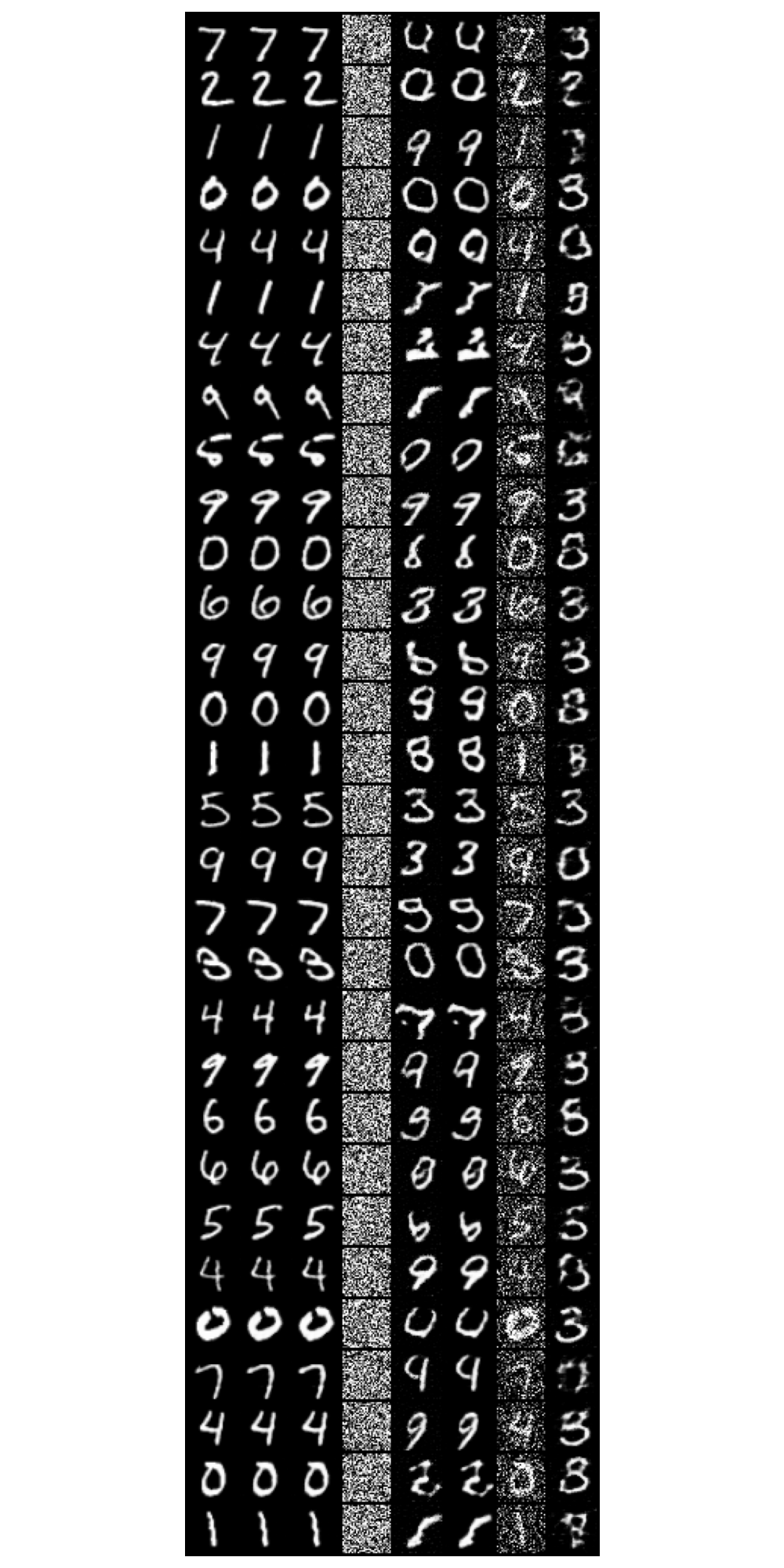}
        \label{fig:subfig1}
    \end{subfigure}
    \begin{subfigure}[b]{0.22\textwidth}
        \centering
        \includegraphics[height=9cm, trim={0pt 0pt 365pt 0pt}, clip]{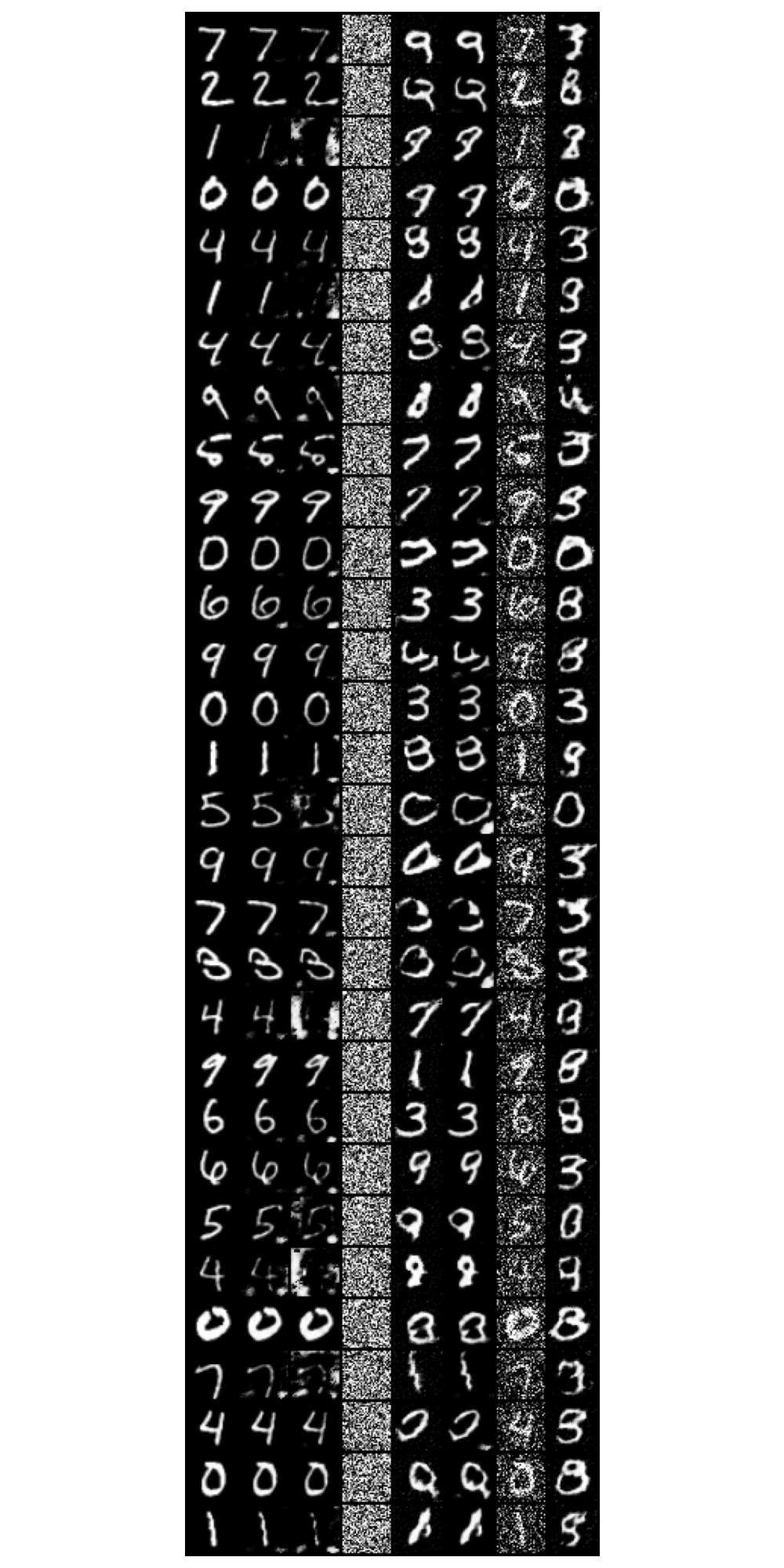}
        \label{fig:subfig2}
    \end{subfigure}    
    \begin{subfigure}[b]{0.22\textwidth}
        \centering
        \includegraphics[height=9cm, trim={0pt 0pt 365pt 0pt}, clip]{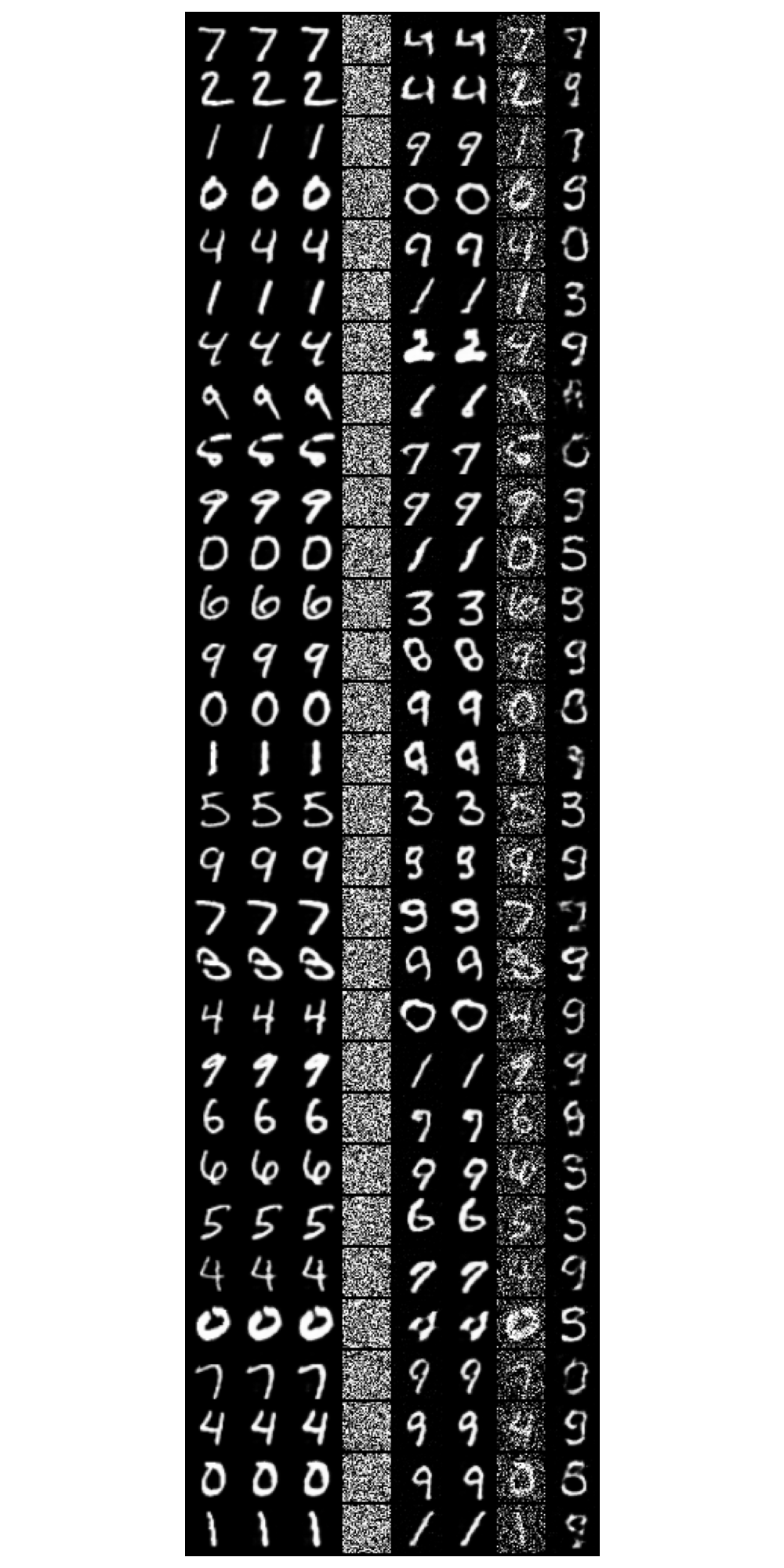}
        \label{fig:subfig2}
    \end{subfigure}    
  \caption{\small{Reconstruction, generation over 3 seeds (one seed per sub-figure) for NAIGN (top) and IGN (bottom). In each sub-figure, the first column depicts images from the test set. Columns 2-3 are the first and second applications of the model on the real images. Columns 3-5 depict noise sampled from the source distribution and the first and second applications of the model, respectively. }}
    \label{fig:appgen}
\end{figure*}

\newpage
\begin{figure*}[b!]
    \centering
    \begin{subfigure}[b]{0.45\textwidth}
        \centering
        \includegraphics[height=10cm]{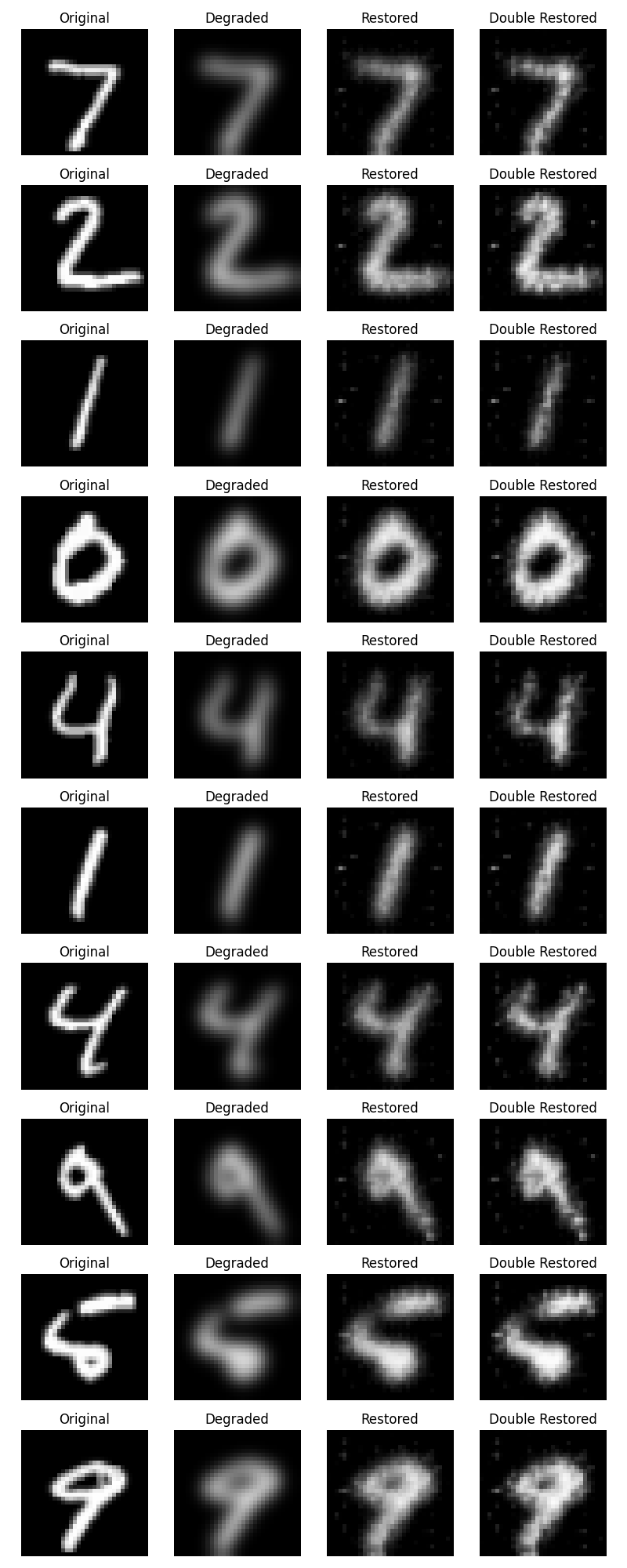}
        \caption{Blur}
        \label{fig:subfig1}
    \end{subfigure}
    \begin{subfigure}[b]{0.45\textwidth}
        \centering
        \includegraphics[height=10cm]{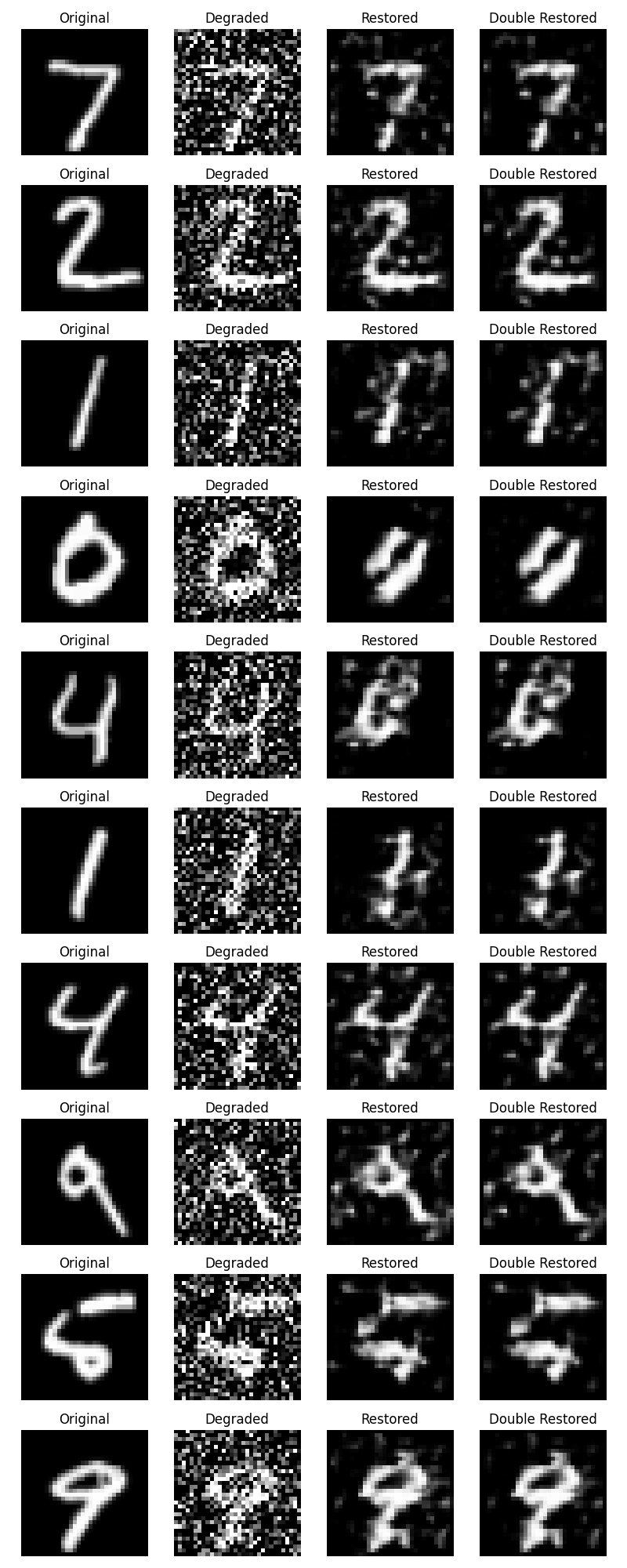}
        \caption{Gaussian Noise}
        \label{fig:subfig2}
    \end{subfigure}    
    \begin{subfigure}[b]{0.45\textwidth}
        \centering
        \includegraphics[height=10cm]{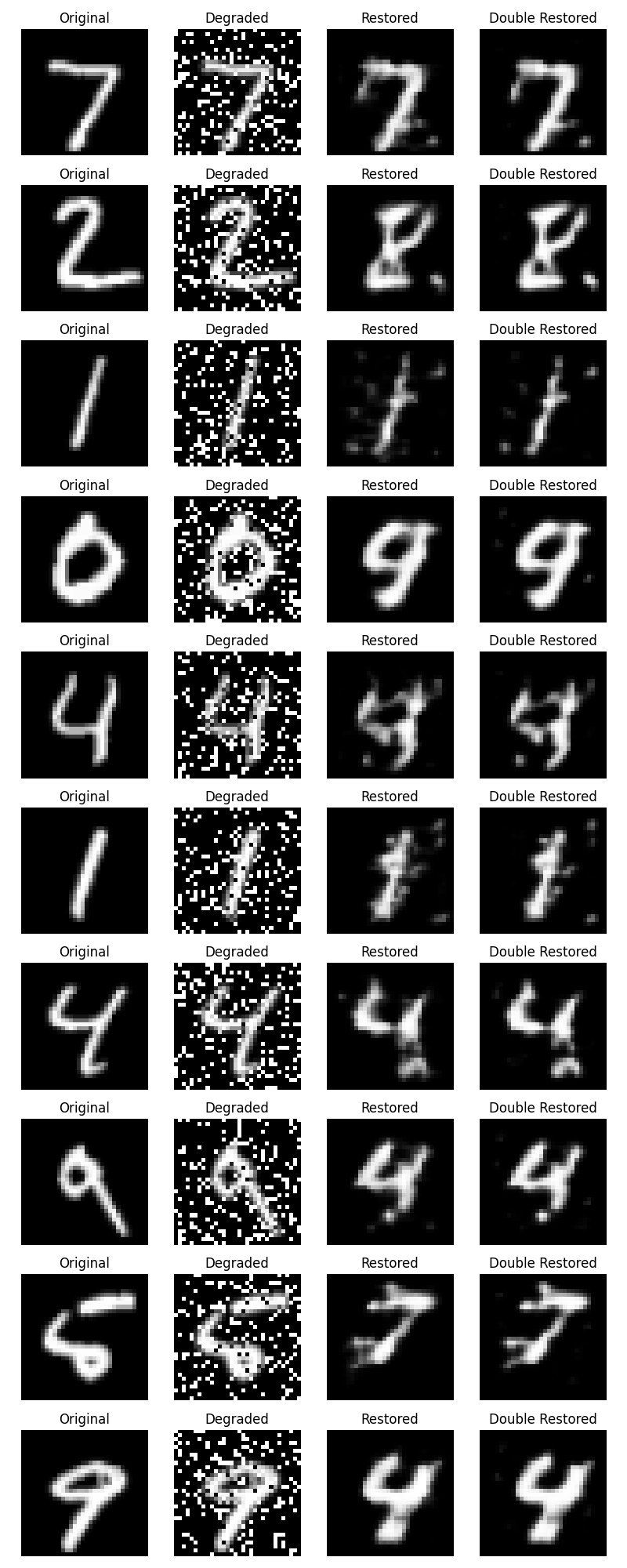}
        \caption{Salt\&Pepper}
        \label{fig:subfig3}
    \end{subfigure}
    \begin{subfigure}[b]{0.45\textwidth}
        \centering
        \includegraphics[height=10cm]{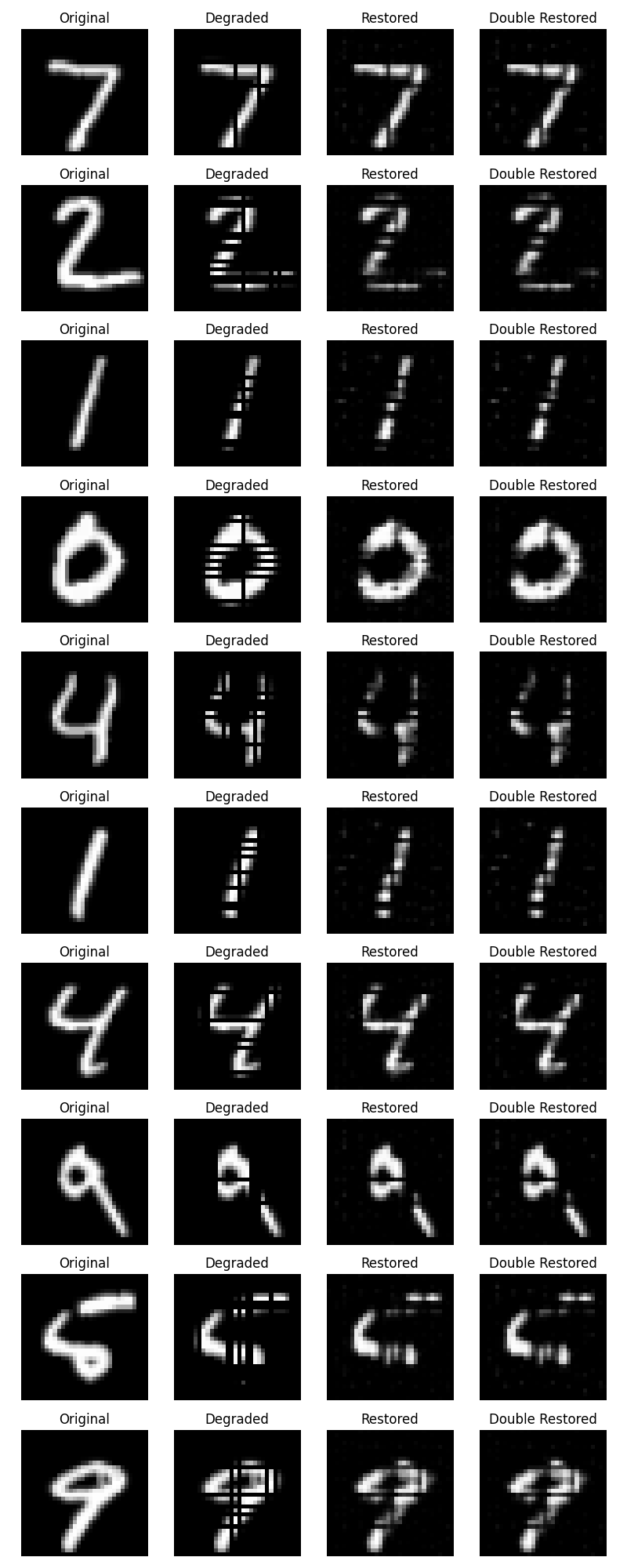}
        \caption{LinesRows}
        \label{fig:subfig4}
    \end{subfigure}
    
    \caption{Restoration for NAIGN.}
    \label{fig:apprest1}
\end{figure*}

\newpage
\begin{figure*}[b!]
    \centering
    \begin{subfigure}[b]{0.45\textwidth}
        \centering
        \includegraphics[height=10cm]{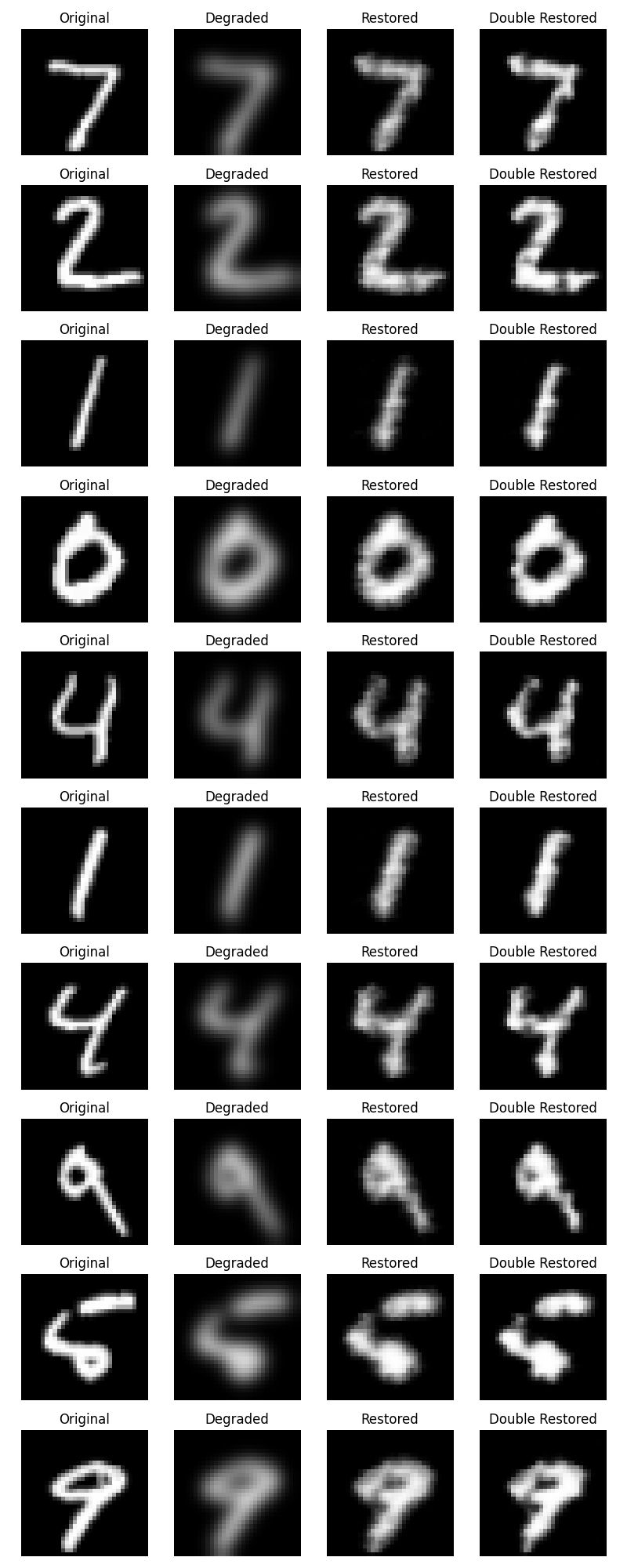}
        \caption{Blur}
        \label{fig:subfig1}
    \end{subfigure}
    \begin{subfigure}[b]{0.45\textwidth}
        \centering
        \includegraphics[height=10cm]{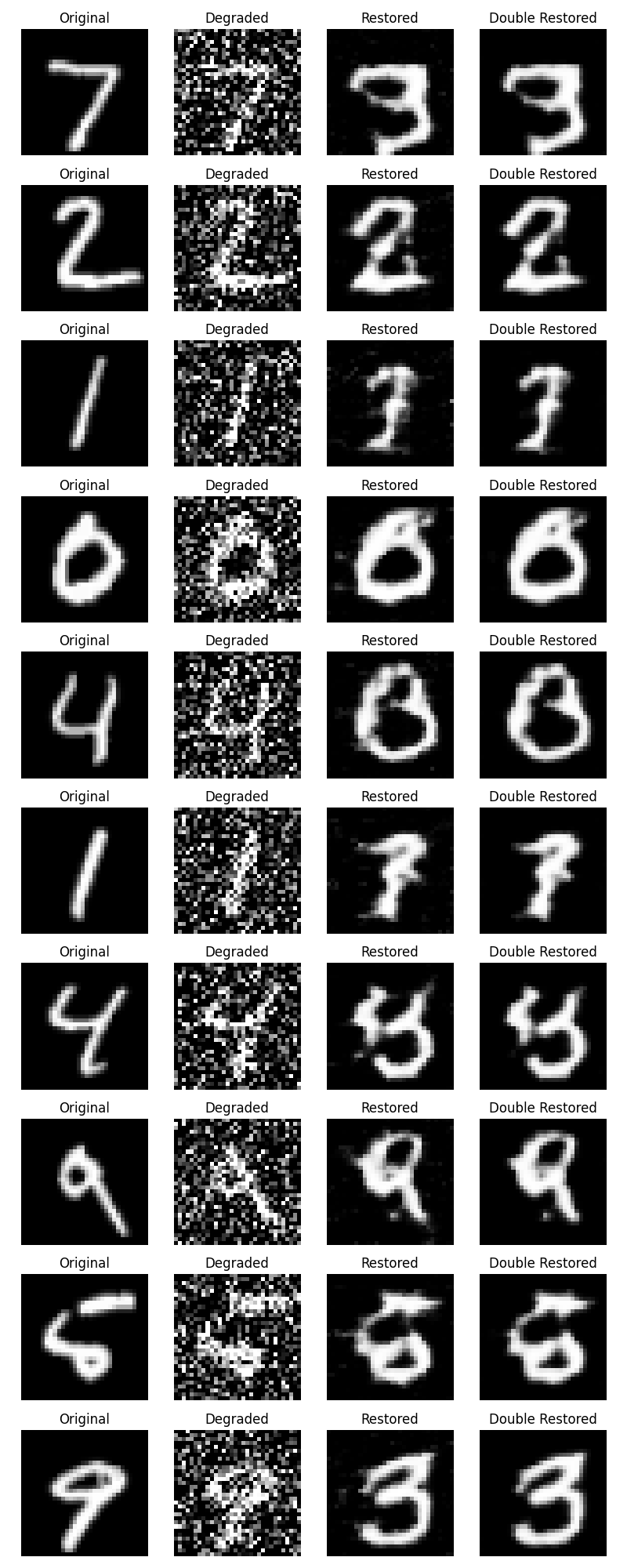}
        \caption{Gaussian Noise}
        \label{fig:subfig2}
    \end{subfigure}    
    \begin{subfigure}[b]{0.45\textwidth}
        \centering
        \includegraphics[height=10cm]{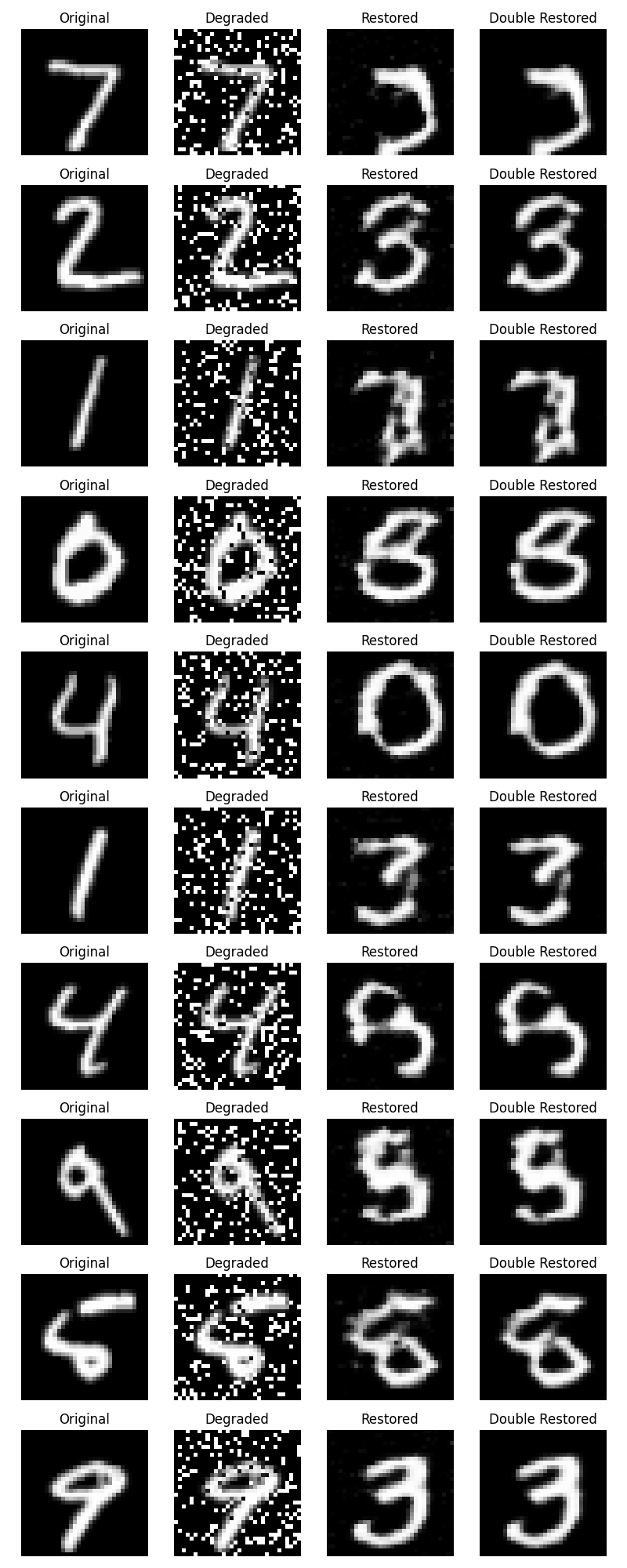}
        \caption{Salt\&Pepper}
        \label{fig:subfig3}
    \end{subfigure}
    \begin{subfigure}[b]{0.45\textwidth}
        \centering
        \includegraphics[height=10cm]{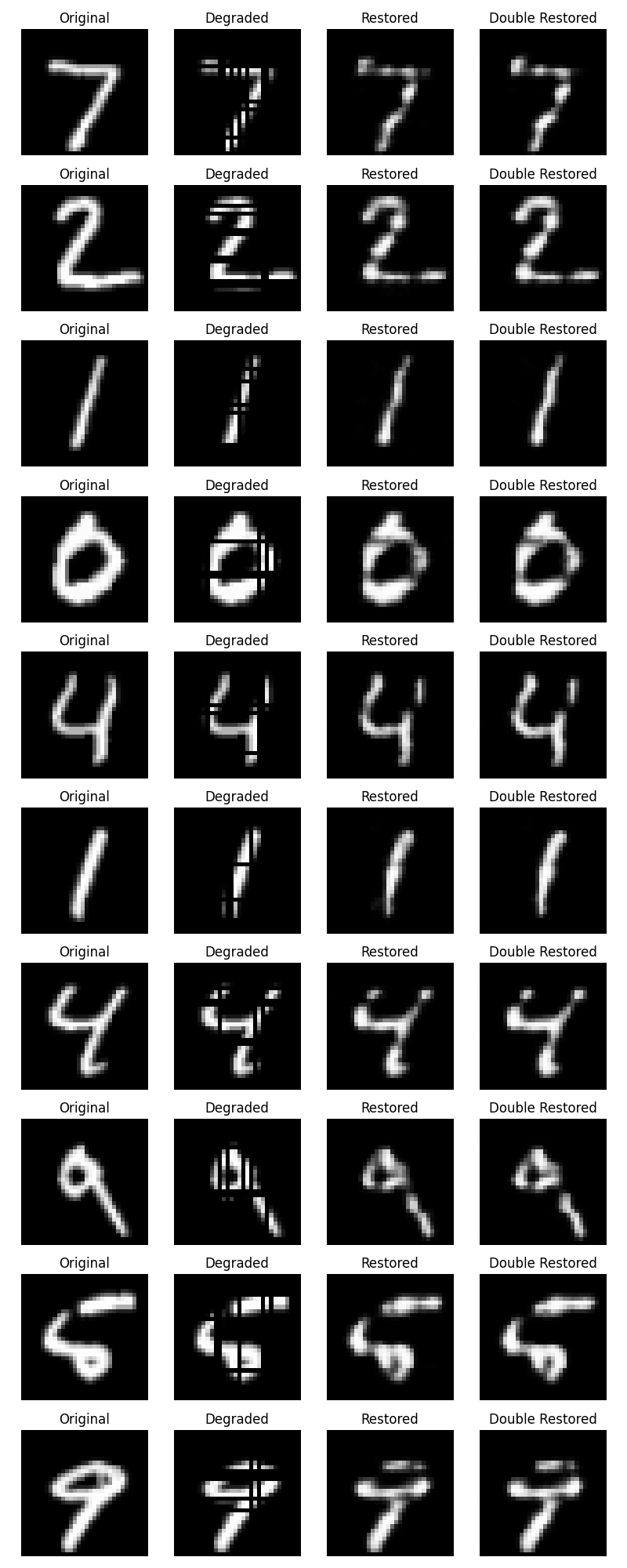}
        \caption{LinesRows}
        \label{fig:subfig4}
    \end{subfigure}
    
    \caption{Restoration for IGN.}
    \label{fig:apprest2}
\end{figure*}
